\documentclass[10pt,twocolumn,letterpaper]{article}

\usepackage{cvpr}      

\definecolor{cvprblue}{rgb}{0.21,0.49,0.74}
\usepackage[pagebackref,breaklinks,colorlinks,allcolors=cvprblue]{hyperref}
\usepackage{graphicx}
\usepackage{subcaption}
\usepackage{amsmath}
\usepackage{tikz}
\usepackage{array}
\usepackage{caption}
\def\paperID{6} 
\def\confName{CVPR}
\def\confYear{2025}
\def\X{\mathbf{X}}
\title{Detecting Localized Deepfake Manipulations Using Action Unit-Guided Video Representations}

\author{
Tharun Anand \quad Siva Sankar \quad Pravin Nair \\
Indian Institute of Technology Madras \\
{\tt\small ed20b068@smail.iitm.ac.in, ch20b103@smail.iitm.ac.in, pravinnair@ee.iitm.ac.in}
}

\begin{document}
\maketitle

\begin{abstract}
With rapid advancements in generative modeling, deepfake techniques are  increasingly narrowing the gap between real and synthetic videos, raising serious privacy and security concerns. Beyond traditional face swapping and reenactment, an emerging trend in recent state-of-the-art deepfake generation methods $(2021$-$2024)$ involves localized edits such as subtle manipulations of specific facial features like raising eyebrows, altering eye shapes, or modifying mouth expressions. These fine-grained manipulations pose a significant challenge for existing detection models, which struggle to capture such localized variations. To the best of our knowledge, this work presents the first detection approach explicitly designed to generalize to localized edits in deepfake videos by leveraging spatiotemporal representations guided by facial action units. Our method leverages a cross-attention-based fusion of representations learned from pretext tasks like random masking and action unit detection, to create an embedding that effectively encodes subtle, localized changes. Comprehensive evaluations across multiple deepfake generation methods demonstrate that our approach, despite being trained solely on the traditional FF+ dataset, sets a new benchmark in detecting recent deepfake-generated videos with fine-grained local edits, achieving a $20\%$ improvement in accuracy over current state-of-the-art detection methods. Additionally, our method delivers competitive performance on standard datasets, highlighting its robustness and generalization across diverse types of local and global forgeries.  
\end{abstract}
\vspace{-8mm}  

\section{Introduction}
\label{sec:intro}

Deepfake generation techniques \cite{zhang2022deepfake,pei2024deepfake} are advancing rapidly, driven by sophisticated editing capabilities enabled by state-of-the-art generative models. In particular, foundational technologies, such as Generative Adversarial Networks (GANs) \cite{goodfellow2020generative} and diffusion models \cite{ho2020denoising} have significantly enhanced the quality of deepfake videos, facilitating diverse manipulation techniques like face swapping  (exchanging identities between individuals) \cite{shiohara2023blendface,han2025face}, face reenactment (transferring movements and poses from one face to another) \cite{hsu2022dual,bounareli2023hyperreenact}, and localized edits (modifying specific facial attributes to change expressions or context) \cite{yao2021latent,tzaban2022stitch,kim2023diffusion}. While these technologies have beneficial applications, their unethical use raises serious privacy concerns over the spread of falsified media content \cite{pantserev2020malicious,karnouskos2020artificial}. The growing prevalence of such videos underscores the critical need for robust and generalizable detection methods to protect against malicious uses of deepfake technology.
\begin{table}
    \centering
    \begin{tabular}{>{\centering\arraybackslash}m{0.13\textwidth} @{\hskip 2pt} 
                     >{\centering\arraybackslash}m{0.13\textwidth}
                     @{\hskip 4pt} >{\centering\arraybackslash}m{0.16\textwidth}}
        \includegraphics[width=\linewidth]{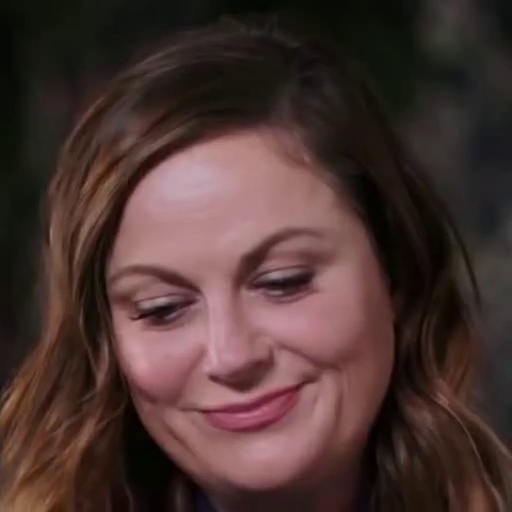} & 
        \includegraphics[width=\linewidth]{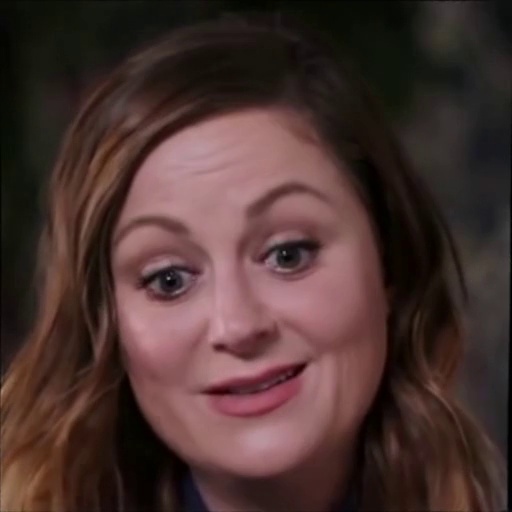} &  
        \includegraphics[width=\linewidth]{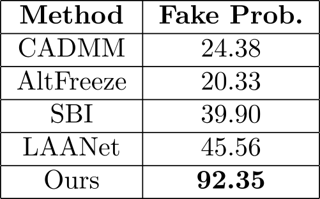} \\        
        \includegraphics[width=\linewidth]{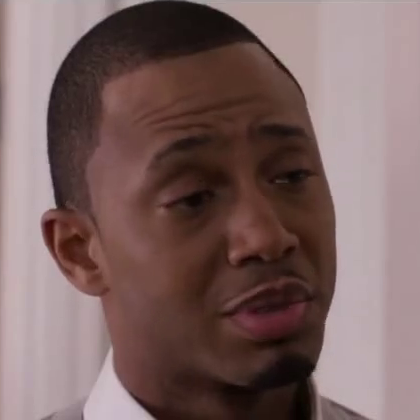} & 
        \includegraphics[width=\linewidth]{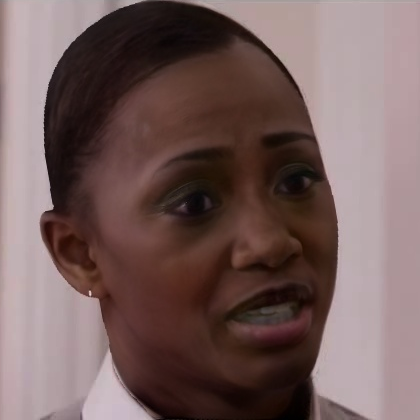} &  
        \includegraphics[width=\linewidth]{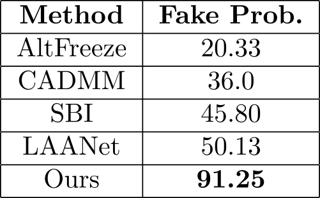} \\
        \includegraphics[width=\linewidth]{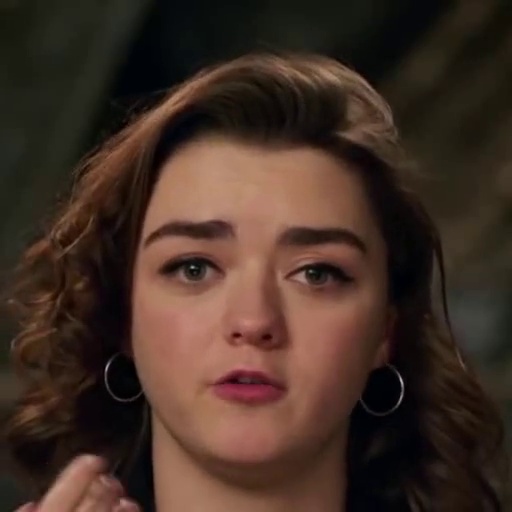} & 
        \includegraphics[width=\linewidth]{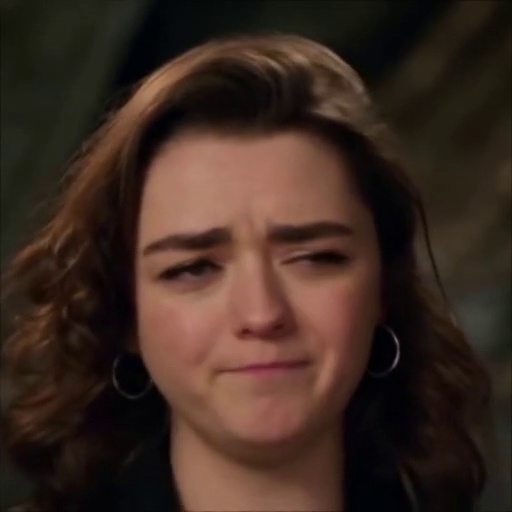} & 
        \includegraphics[width=\linewidth]{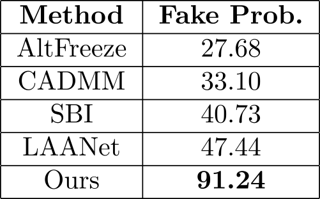} \\        
        \textbf{\footnotesize{Original}} & \textbf{ \footnotesize{Fake}} & \textbf{ \footnotesize{Probability score}} \\
    \end{tabular}
\captionof{figure}{\textbf{Locally Edited Deepfakes Detection:} A real video is manipulated to produce fake videos with subtle hard-to-detect  edits - \textbf{raised eyebrows, gender modification, expression change to disgust} (single frame shown for illustration). Our method achieves significantly higher probability scores over top methods, effectively detecting  these fine-grained edits with high confidence.}
\label{fig:motexample}
\end{table}

Although existing detection methods have shown reasonable success in identifying deepfakes involving face swaps and reenactments, they fall short when it comes to detecting localized manipulations of specific facial attributes. Unlike traditional deepfakes that rely on full-face alterations, these emerging techniques focus on nuanced edits, such as subtle changes to the shape of the mouth, nose, or eyes, or adjustments to micro-expressions like smiles or frowns. Such targeted manipulations produce synthetic content that is nearly indistinguishable from authentic media, posing a formidable challenge for current detection models, which are typically optimized for detecting broader, more apparent alterations. This shift toward high-fidelity, localized edits introduces a significant gap in current detection capabilities, leaving existing methods vulnerable to these next-generation deepfakes. Refer to  Fig.~\ref{fig:motexample} for visual examples of localized manipulations. 

\begin{figure*}
\centering
\includegraphics[width=1.0\textwidth]{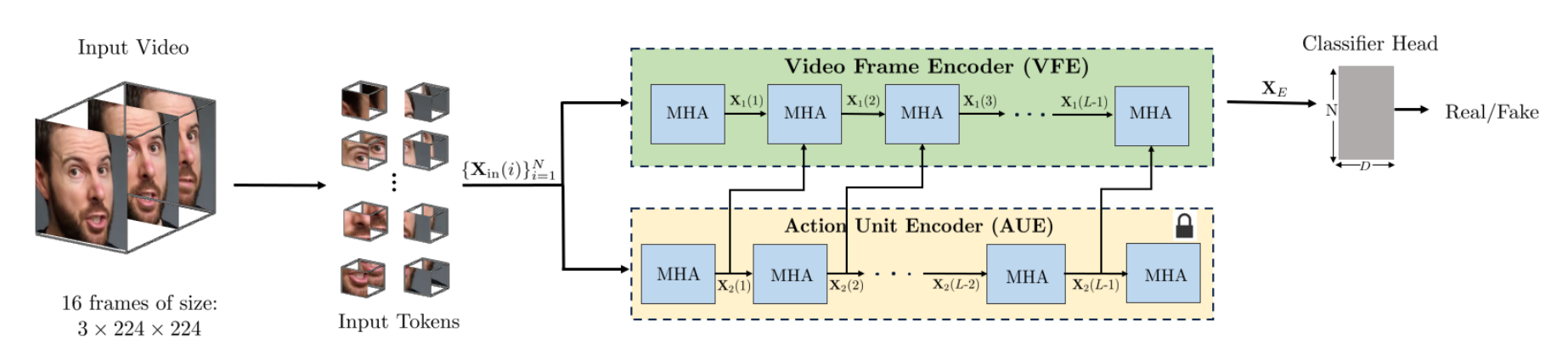} 
\caption{\textbf{Proposed Method}: The input video is processed using a face detection algorithm to extract equally spaced face-centered frames. These frames are divided into $N$ tubular patches, which are fed into a novel encoder, obtained by fusing latent representations obtained from pretrained pretext tasks, to generate latent vector $\X_E$. The encoded latent vector $\X_E$ is then passed through a classification head to detect the video as real or fake.}
\label{Propfig}
\end{figure*}

Addressing this challenge requires a unified detection framework capable of accurately identifying not only face swaps and reenactments, but also fine-grained, localized manipulations. To meet this need, we propose a generalization-focused detection framework trained solely on the widely used FaceForensics++ (FF++) dataset \cite{rossler2019faceforensics}. Although FF++ primarily includes videos generated by earlier deepfake methods focused on face swapping and reenactment, our method demonstrates the ability to generalize from this dataset to detect advanced, high-fidelity localized edits. This generalization capability hinges on learning optimal spatio-temporal representations that capture subtle, frame-level variations, enabling consistent detection across a wide range of manipulation types.

Our approach leverages carefully designed pretext tasks to drive representation learning. Specifically, we introduce a novel action unit-guided spatio-temporal  framework that combines Facial Action Units (AUs) detection and masking-based pretext tasks to learn robust neural representations. Defined by the Facial Action Coding System (FACS) \cite{ekman1978facial}, AUs collectively describe a wide range of expressions and micro-expressions, where each AU corresponds to a unique movement, such as eyebrow raises, eyelid movements, lip pulls, etc.  By using AU-based features to guide video representations learned through Masked Autoencoders (MAE) \cite{tong2022videomae}, our method effectively captures localized changes crucial for detecting subtle facial edits. This approach enables us to construct a unified latent representation that encodes both localized edits and broader alterations in face-centered videos, providing a comprehensive and adaptable solution for deepfake detection. 

In this regard, our key contributions are as follows:
\begin{enumerate}
\item We propose a novel framework for learning robust spatio-temporal representations of videos, guided by Facial Action Unit (AU) embeddings. Through a cross-attention mechanism, our approach fuses frame-level features with AU-derived embeddings, capturing subtle, localized manipulations with high sensitivity. 

\item Trained solely on standard deepfake FF+ dataset, our model achieves strong generalization to advanced deepfake techniques that involve high-fidelity, localized edits, where our model achieves a $20$\% improvement in detection accuracy over the latest state-of-the-art deepfake detection methods. 

\item To the best of our knowledge, this is the first approach specifically designed to detect localized edits in deepfake videos. Our AU-driven representation learning is not only effective for localized edits but also competitive on traditional deepfake datasets, offering a scalable, future-proof solution for diverse deepfake challenges.
\end{enumerate}       

\section{Related Work}

Early deepfake detection methods focussed on image-level detectors that identify spatial artifacts within individual frames  \cite{afchar2018mesonet,bayar2016deep,chai2020makes,fu2022m3l,hsu2018learning,rahmouni2017distinguishing}. These methods primarily employed variations of Convolutional  Neural Networks (CNNs) \cite{krizhevsky2012imagenet}, with networks like EfficientNet \cite{tan2019efficientnet} and XceptionNet \cite{ashok2023deepfake} becoming standard baselines due to their effectiveness. To enhance detection, researchers introduced techniques that operate in the frequency domain, detecting subtle artifacts often missed in the RGB domain \cite{jeong2022bihpf,jeong2022frepgan,masi2020two,qian2020thinking}. Additional approaches targeted face blending artifacts \cite{chen2022self,li2020face} and artifacts due to resolution discrepancies between source and target video  \cite{li2018exposing}, further improving frame-level detection. However, these methods were limited by their inability to capture temporal inconsistencies present in deepfake videos.

To address this limitation, video-level detectors were developed, leveraging temporal information across multiple frames \cite{amerini2019deepfake,cozzolino2021id,de2020deepfake,haliassos2021lips,li2018ictu,mittal2020emotions,sabir2019recurrent}. CNN-based recurrent models \cite{sabir2019recurrent} incorporated recurrent units after CNNs to capture temporal dynamics, while other methods directly learned spatiotemporal features \cite{amerini2019deepfake,de2020deepfake,zheng2021exploring}.  Domain-specific insights, such as facial action units \cite{bai2023aunet,agarwal2019protecting}, lip motion \cite{haliassos2021lips}, and identity inconsistency \cite{dong2022protecting,cozzolino2021id}, further improved detection models by exploiting unnatural temporal variation in deepfake videos. Recently, advancements in deepfake detection involve Vision Transformers (ViT)  \cite{dong2022protecting,coccomini2022combining,khan2022hybrid,bai2023aunet}, which focus on spatio-temporal regions within frames to detect low-level perturbations in manipulated videos.

With the rising threat of deepfakes in recent years, current research efforts focus on developing generalizable deepfake detection methods that capture facial cues robustly across diverse forgery patterns \cite{nguyen2024laa,shiohara2023blendface,wang2023altfreezing,lin2024preserving,zhou2023instance}. Most of the recent methods are fairly generalizable in detecting global manipulations like face swapping \cite{shiohara2023blendface,han2025face}, face reenactment \cite{hsu2022dual,bounareli2023hyperreenact} etc. 
However, with recent advances in deepfake generation, a new challenging form of deepfakes has emerged where a person's facial feature or expression can be minutely changed, which can alter the context of the video \cite{yao2021latent,tzaban2022stitch,kim2023diffusion}. To the best of our knowledge, prior to our work, the generalizability of current state-of-the-art detection methods to recent deepfake generation techniques involving localized manipulations  \cite{yao2021latent,tzaban2022stitch,kim2023diffusion}, has remained unexplored. 

\section{Proposed Method}
\label{sec:method}
The proposed method is illustrated in Fig.~\ref{Propfig}. Given a video, we first perform face detection on these frames and then sample equally spaced frames. Following the tokenization approach in Video Masked Autoencoders (VideoMAE) \cite{tong2022videomae}, these set of frames is then divided into  $N$ tubular (3D) blocks, denoted by \(\{\X_{\text{in}}(i)\}_{i=1}^N\), each of size \(T \times P \times P\), where every token captures localized  spatial and temporal information. The tokenization process is refined using a patch embedding layer, followed by the addition of positional encodings to preserve spatial-temporal relationships.

The complete token set is then passed through the proposed encoder, producing a latent representation \(\mathbf{X}_{\text{E}}\) of dimensions \(N \times D\), where \(D \) denotes the token dimensionality across all tubular blocks. This latent representation is then processed by a classification head to determine if the video is real or fake. Although the framework is compatible with various encoder architectures, achieving a truly generalizable latent representation that can robustly distinguish between real and synthetic content, even in the case of localized manipulations, remains a significant challenge.

To address this, we next introduce the design of our novel encoder, tailored specifically to generate rich, discriminative features optimized for detecting deepfake videos with localized  edits. Our encoder construction is based on learning robust latent embeddings through two complementary pretext tasks, followed by an effective cross-attention based fusion of these embeddings. Next, we provide a comprehensive analysis of the designed pretext tasks and the architectural formulation of the proposed encoder.

\subsection{Learning Representations using Pretext Tasks}

\begin{figure*}
\centering
\includegraphics[width=\textwidth]{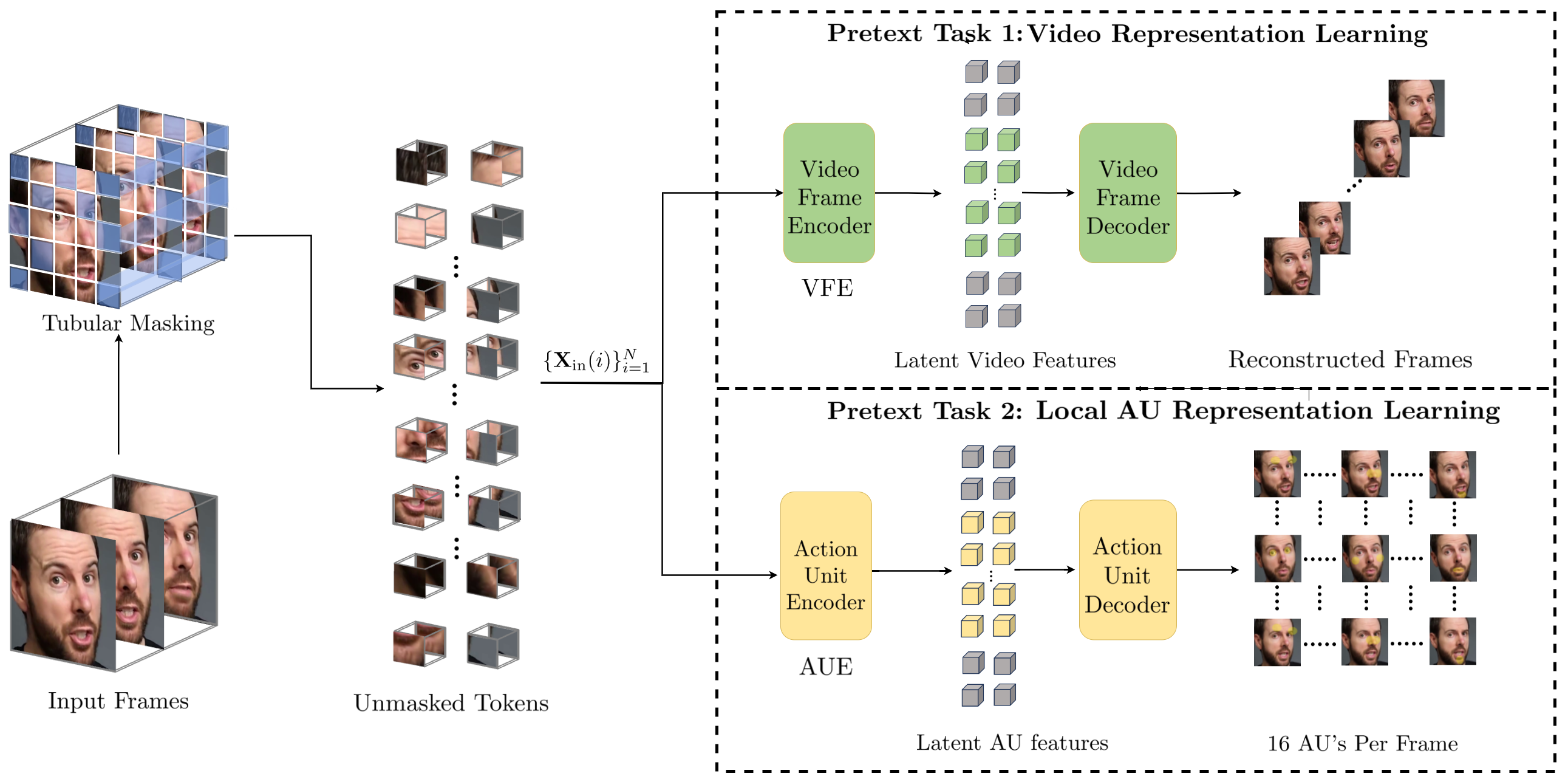} 
\caption{\textbf{Pretext Tasks Training}: Video-derived tubular tokens are first processed to form learnable embeddings. Some tokens are randomly masked, and the visible tokens are refined using an encoder. For training, the encoded latent representation is appended with placeholder tokens for masked positions (shown in gray) and passed through the decoder to reconstruct task-specific targets. For masking, the target is same as input face-centered frames and for AU detection, the target to be reconstructed is $16$ action unit maps per frame.}
\label{Pretext}
\end{figure*}

The overall procedure for training our pretext tasks is illustrated in Fig.~\ref{Pretext}. The tubular tokens \(\{\mathbf{X}_{\text{in}}(i)\}_{i=1}^N\) are initially projected to form learnable embeddings, subsequently reshaping them into token embeddings of size \({N \times D}\). We apply a structured masking strategy to occlude a fraction of tubular tokens, ensuring that only \(M\) tokens remain visible. This process partitions the latent representation into a visible subset of dimension \({M \times D}\) and a masked subset of dimension \({(N-M) \times D}\). Such a masking strategy enhances the difficulty of the reconstruction task, encouraging the encoder to learn robust and detailed spatiotemporal structures by focusing on limited visible information.

For our pretext tasks, we adopt the encoder-decoder architecture from VideoMAE \cite{tong2022videomae}. As a preprocessing step, the input video is first tokenized using a patch embedding layer, followed by the addition of positional encodings to retain spatial-temporal information. We choose to mask the tubular tokens by masking the corresponding positionally encoded tokens. Then after masking, the visible tokens are processed by an encoder composed of $L$ multi-head self-attention blocks. The resulting latent embeddings are then fed into a decoder, which reconstructs the task-specific target by leveraging the partially visible information from the encoder and the placeholder learnable tokens for the masked regions (indicated in gray). The decoder architecture mirrors the encoder, preserving input dimensions throughout the process. Both encoder and decoder weights, as well as the projections for embeddings, are optimized using task-specific loss functions to ensure robust and detailed representation learning. 

\noindent \textbf{\underline{Unmasked Video Frame Reconstruction}}:  
The first pretext task involves reconstructing masked regions within input video frames, allowing the model to interpolate masked areas by leveraging dependencies within frames. This technique was initially proposed for representation learning in images \cite{pathak2016context} and eventually extended to videos \cite{he2022masked,wang2022bevt,tong2022videomae}. 
By training the model to reconstruct the masked areas, we enable the model to learn rich spatio-temporal relationships and develop an implicit understanding of scene dynamics. 

The reconstruction target is the original video frames, with the decoder producing a sequence of face-centered frames that closely approximates the input. This video frame reconstruction task  enables the encoder, denoted as Video Frame Encoder (VFE), to produce representations that represent the visible tokens, focusing on reconstructing meaningful spatiotemporal features. Although these latent embeddings can be directly applied for the downstream task of deepfake detection, our results in Table \ref{tab:ablation_e2_new} indicate that these learned features struggle to consistently differentiate between real and manipulated content, especially in the case of localized manipulations.  Consequently, it becomes necessary to refine the representations learned by VFE to address the generalization challenge in deepfake detection.

\noindent \textbf{\underline{AU Detection}}: To enhance the representations learned from the masking, we introduce another pretext task centered on local AU detection. Leveraging an encoder-decoder structure similar to the masking model architecture, this pretext task is designed to learn fine-grained local facial dynamics, capturing subtle expressions and muscle movements. To construct latent representations for AU detection, we follow a procedure similar to the random masking pretext task, with the only change that the decoder is architectured for the reconstruction target to be AU detection output. In particular, for a frame of dimensions, $3 \times H \times W$ and a chosen set of $F$ AUs, the reconstruction target is an $F \times 3 \times H \times W$ label map, where each channel corresponds to a detected AU, capturing finer facial cues for robust feature extraction. 

The resulting AU-based latent representations from the encoder serve as a complementary set of features to video representation features obtained from masking pretext task, which we will effectively use to enrich the spatio-temporal embeddings learned from masking.  These latent representations for AU detection can also be applied directly for downstream deepfake detection. However, similar to the latent embeddings from masking i.e., the representations output by VFE, our empirical analysis indicates that the AUE's latent representations lack sufficient generalization across diverse, locally edited deepfake videos, as demonstrated in Table \ref{tab:cross_dataset_auc} in Section \ref{sec:experiments}. In the following section, we use these AU-based latent representations as conditioning vectors to significantly enhance the robustness and generalization of the latent representations obtained through masking. 

\subsection{Fused Latent Representation}

We next discuss the construction the final fused encoder. The tubular tokens \(\{\mathbf{X}_{\text{in}}(i)\}_{i=1}^N\) are processed through patch embedding layers corresponding to  both our pretrained encoders, VFE and AUE. Let the resulting two latent vectors be denoted \(\mathbf{X}_1\) and \(\mathbf{X}_2\), corresponding to masking and AU detection pretext tasks respectively.

As shown in Fig.~\ref{Propfig}, in the fused encoder, the Multi-head attention blocks from layers $2$ to \(L\) in VFE are conditioned using cross-attention,  i.e. the query embeddings are derived from the corresponding layers of AUE. Specifically, let the output of every $i\mbox{-}$th layer in VFE be denoted by $\X_1(i)$, and that of AUE be denoted by $\X_2(i)$. 
Then, each block's output in VFE is computed as, 
\begin{equation*}
\mathbf{X}_1(i) = \text{Concat}(\text{head}_1, \text{head}_2, \dots, \text{head}_H) \mathbf{W}_O,
\end{equation*}
where each attention head is defined as,
\begin{equation*}
    \scalebox{0.92}{$
    \text{head}_h = \text{softmax} \left( \frac{\mathbf{X}_2(i-1) \mathbf{W}_Q^h (\mathbf{X}_1(i-1) \mathbf{W}_K^h)^T}{\sqrt{d}} \right) \mathbf{X}_1(i-1) \mathbf{W}_V^h, 
    $}
\end{equation*}
with weight matrices \(\mathbf{W}_Q^h, \mathbf{W}_K^h, \mathbf{W}_V^h \in \mathbb{R}^{D \times d}\), \(\mathbf{W}_O \in \mathbb{R}^{D \times D}\), \(d = D / H\), and \(H\) denoting the number of attention heads. This fused cross-attention structure enables the final output latent vector \(\mathbf{X}_1(L) \in \mathbb{R}^{N \times D}\) to serve as the fused representation \(\mathbf{X}_E\) from the encoder.
\subsection{Implementation Details}

\noindent \textbf{Preprocessing and Data Preparation:}  
We preprocess input videos using FaceXZoo \cite{wang2021facex} for face detection, thereby extracting $16$ face-centered frames per video. This results in a sequence of face-centered frames of dimensions $16 \times 3 \times 224 \times 224$. Each sequence is then partitioned into 3D spatiotemporal blocks of size $2 \times 16 \times 16$, where $T=2$ denotes the temporal dimension and $P=16$ represents the spatial patch size, resulting in $N=1568$ tokens per video.

\noindent \textbf{Pretext Task Training:}  
We train the pretext tasks on the CelebV-HQ dataset \cite{zhu2022celebv}, which consists of $35,000$ high-quality facial videos. Following VideoMAE \cite{tong2022videomae}, we employ a $50\%$ random masking strategy during training to learn robust features. The encoder and decoder architectures utilize $L=11$ layers of multi-head attention, with a fixed token embedding dimension of $D=768$. 

For the masked frame reconstruction pretext task, we adopt an $\ell_1$ reconstruction loss, minimizing the difference between ground-truth frames and their reconstructions from masked inputs. For the Action Unit (AU) detection pretext task, we define a set of $F=16$ AUs, capturing fine-grained facial movements across key regions such as the \textit{eyebrows, eyelids, nose, cheeks, lips, and dimples}. We obtain the required ground-truth AU annotations for supervised learning of AU detection using state-of-the-art AU detection techniques \cite{jacob2021facial}. The AU detection model produces a structured output of dimensions $16 \times 3 \times 224 \times 224$, where each channel encodes an individual AU. Our selection of AUs is validated through extensive ablation studies in Section \ref{sec:experiments}. Similar to masked frame reconstruction task, we adopt an $\ell_1$ loss for AU detection, minimizing the difference between ground-truth AU representations and our model output. 

\noindent \textbf{Fine-tuning for Deepfake Detection:}  
For downstream deepfake detection, we fuse the pretrained encoders into a unified representation, as shown in Fig.~\ref{Propfig}, and integrate a classifier. The finetuning is performed on the FaceForensics++ (FF++) dataset \cite{rossler2019faceforensics}, which contains $700$ real and $3,600$ deepfake videos generated using state-of-the-art face swapping and reenactment techniques  \cite{kowalski2024faceswap,thies2016face2face,deepfakes_github,thies2019deferred}. To address class imbalance, we use Focal Loss, a variant of cross-entropy loss \cite{ross2017focal}, ensuring improved learning from hard-to-classify samples.  

\noindent \textbf{Training Setup.}  
Pretext tasks and deepfake detection trainings are conducted separately on a single NVIDIA RTX $4090$ GPU. Additional details on training hyperparameters and setup are provided in the Appendix~\ref{APP_a}.

\section{Experiments}
\label{sec:experiments}

\begin{table*}
    \centering
    \begin{tabular}{>{\centering\arraybackslash}m{0.18\textwidth} @{\hskip 3pt} 
                     @{\hskip 3pt} >{\centering\arraybackslash}m{0.18\textwidth} 
                    @{\hskip 0.5pt} >{\centering\arraybackslash}m{0.18\textwidth} 
                    @{\hskip 0.5pt} >{\centering\arraybackslash}m{0.18\textwidth} @{\hskip 3pt}
                     @{\hskip 3pt} >{\centering\arraybackslash}m{0.22\textwidth}}
        \includegraphics[width=0.95\linewidth]{Figures/Video9/img_9_original.jpg} & 
        \includegraphics[width=0.95\linewidth]{Figures/Video9/img_9_eye_raise.jpg} & 
        \includegraphics[width=0.95\linewidth]{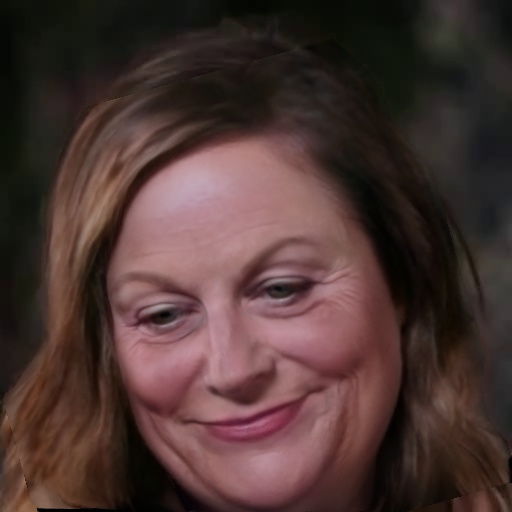} & 
        \includegraphics[width=0.95\linewidth]{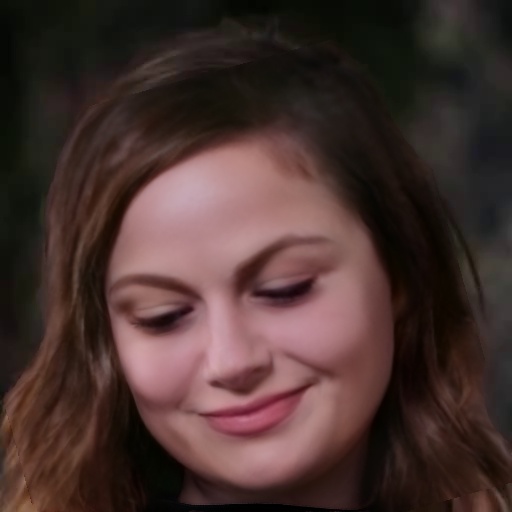} & 
        \textit{\textbf{\footnotesize Eye Raise, Old, Young}}  \includegraphics[width=\linewidth]{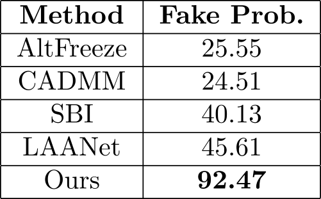} \\        
        \includegraphics[width=0.95\linewidth]{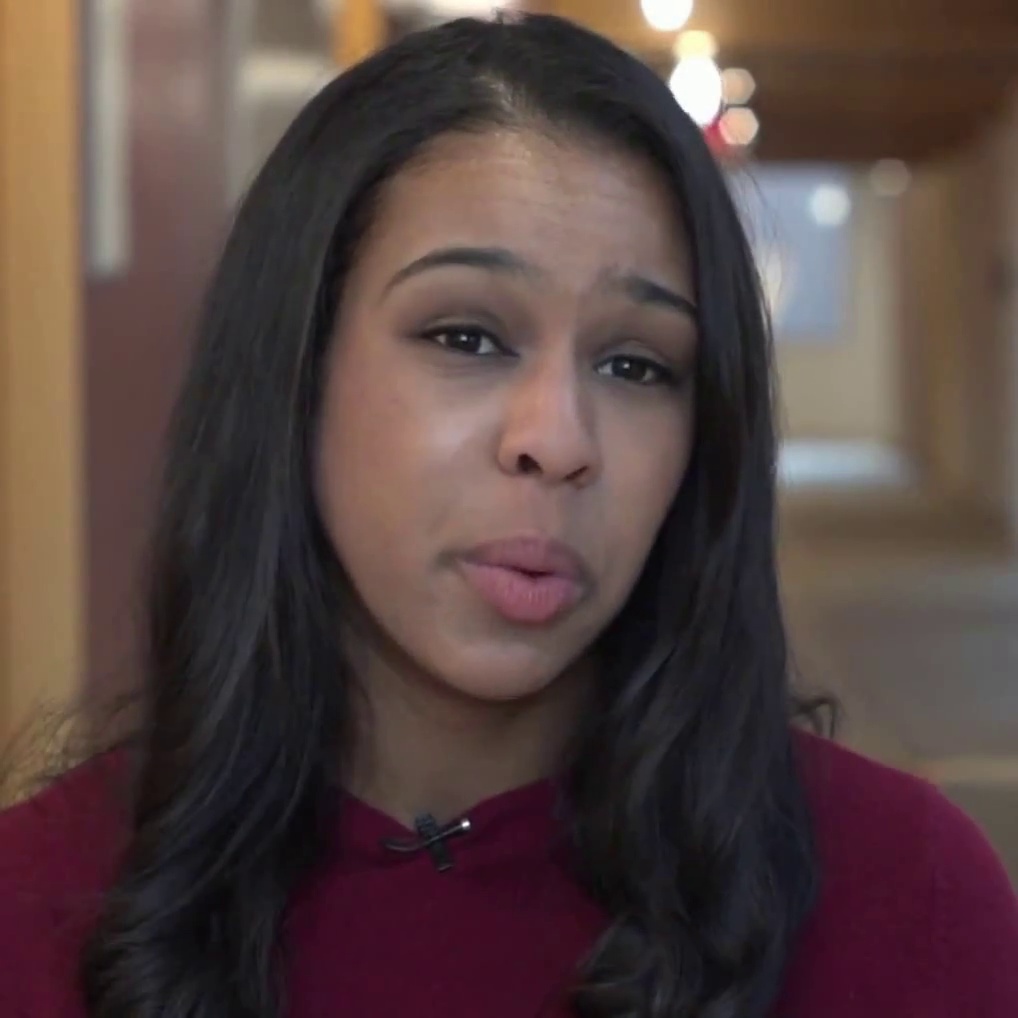} & 
        \includegraphics[width=0.95\linewidth]{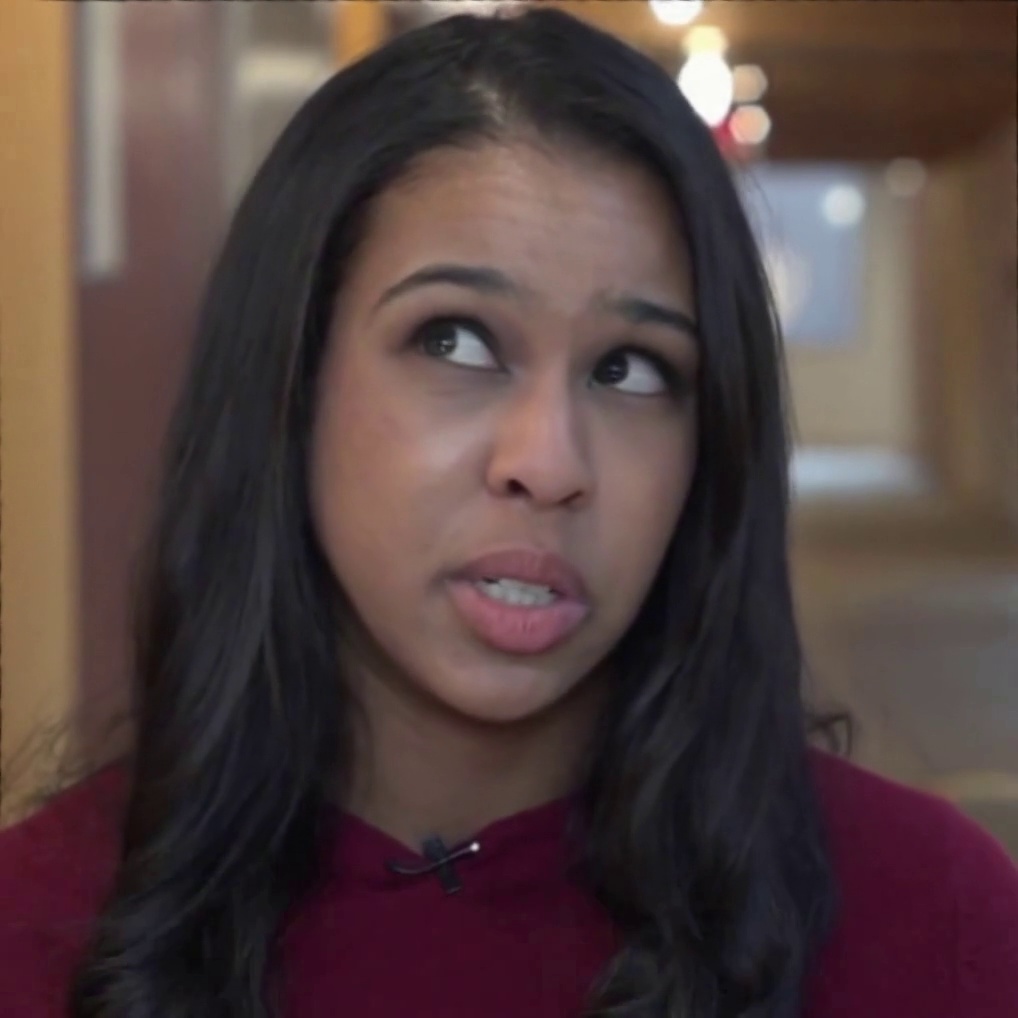} & 
        \includegraphics[width=0.95\linewidth]{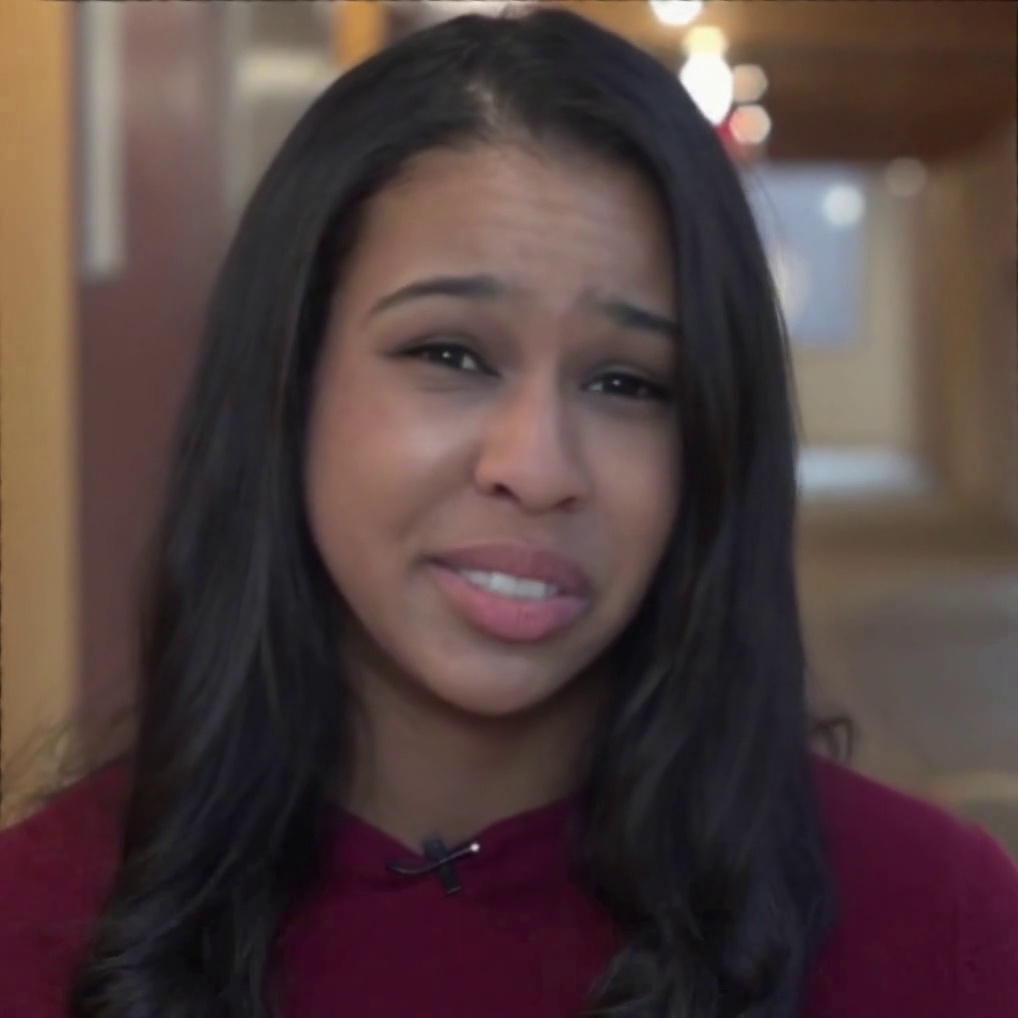} & 
        \includegraphics[width=0.95\linewidth]{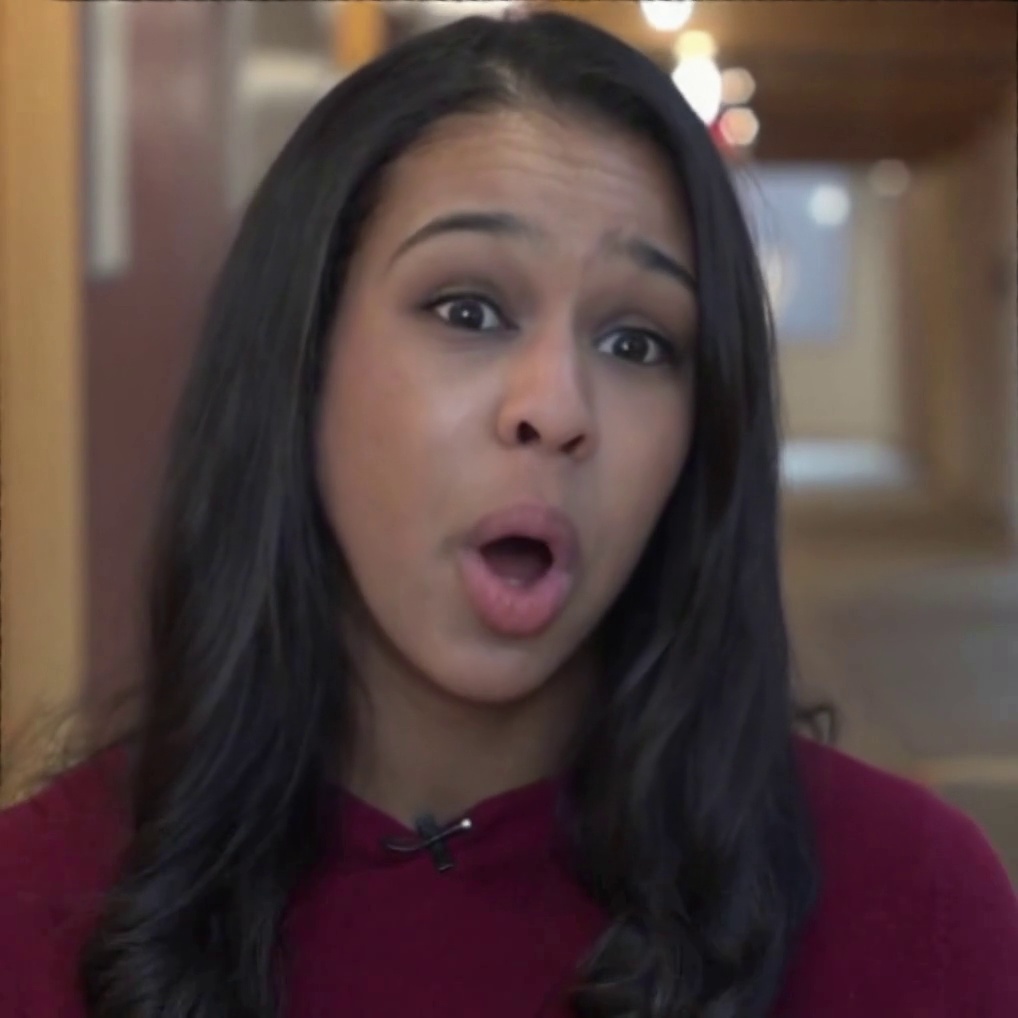} & 
        \textit{\textbf{\footnotesize Eye Gaze, Sad, Shock}} 
        \includegraphics[width=\linewidth]{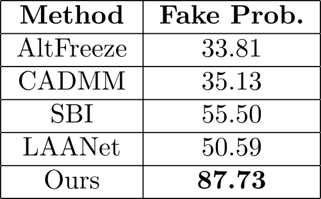} \\        
          \includegraphics[width=0.95\linewidth]{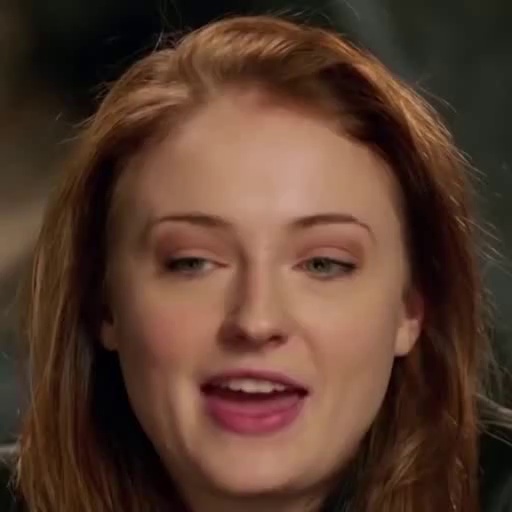} & 
        \includegraphics[width=0.95\linewidth]{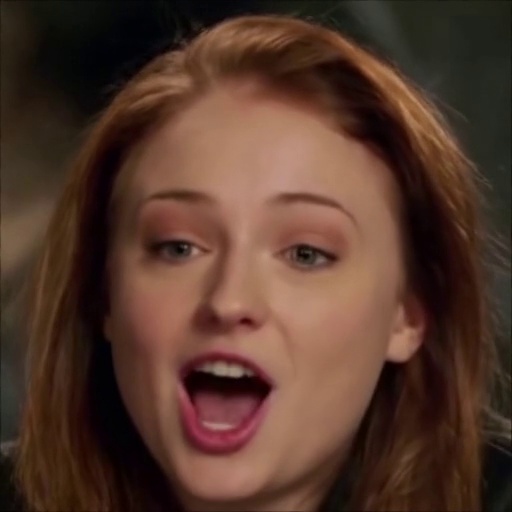} & 
        \includegraphics[width=0.95\linewidth]{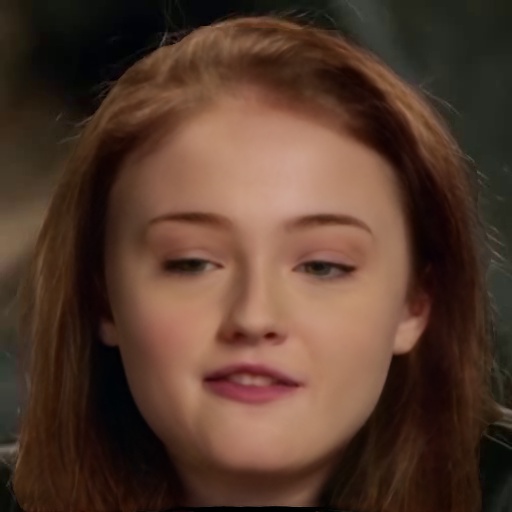} & 
        \includegraphics[width=0.95\linewidth]{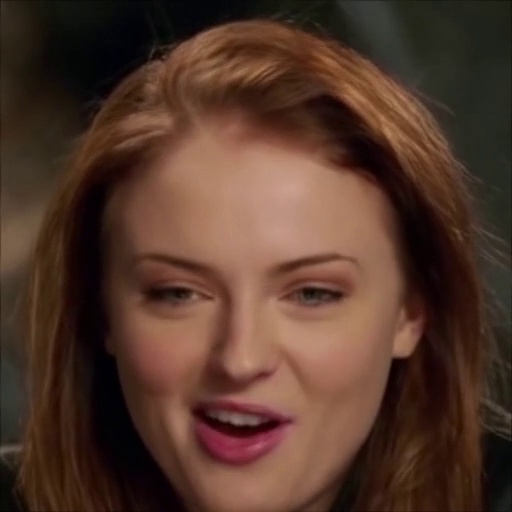} & 
        \textit{\textbf{\footnotesize Shock, Young, Angry}} 
        \includegraphics[width=\linewidth]{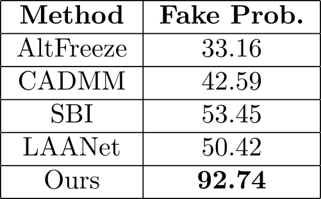} \\        
        \includegraphics[width=0.95\linewidth]{Figures/Video11/img_11_original.jpg} & 
        \includegraphics[width=0.95\linewidth]{Figures/Video11/img_11_disgust.jpg} & 
        \includegraphics[width=0.95\linewidth]{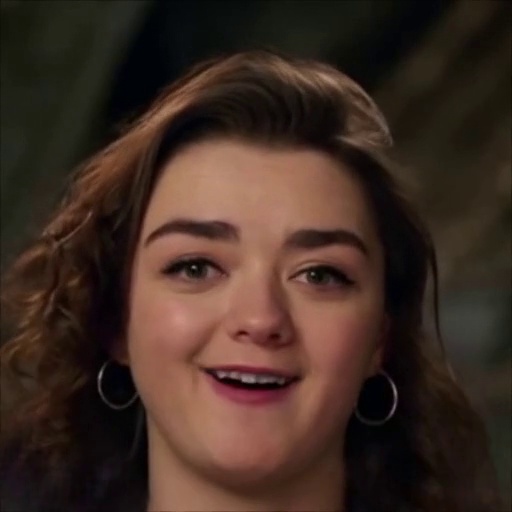} & 
        \includegraphics[width=0.95\linewidth]{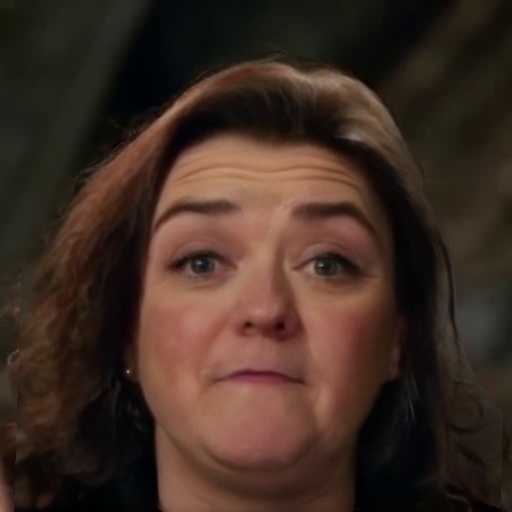} & 
        \textit{\textbf{\footnotesize Disgust, Smile, Old}} 
        \includegraphics[width=\linewidth]{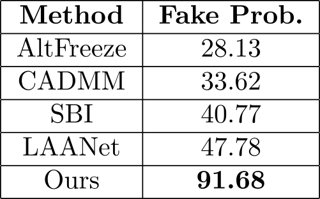} \\        
        \textbf{\footnotesize Original} & \multicolumn{3}{c}{\textbf{\footnotesize Fake Manipulations}} & \textbf{\footnotesize \textit{Edit types} \& Probability score}     
    \end{tabular}
    \captionof{figure}{\textbf{Visual detection comparison for locally manipulated videos:} A real video is subjected to three types of localized manipulations, creating fake videos that are visually indistinguishable from the original. Individual frames for real and fake videos are shown for illustration. For each real sample, we tabulate the fake probability scores averaged for the three edited versions, comparing our method  with state-of-the-art approaches. Despite the subtlety of the edits, our method successfully identifies the fake videos with high probability, unlike latest detection methods, demonstrating superior sensitivity to localized manipulations that are undetectable even to the naked eye.}
    \label{fig:example}
\end{table*}

In this section, we present a comprehensive evaluation of our proposed method, including both quantitative and qualitative comparisons against state-of-the-art deepfake detection methods: FTCN \cite{zheng2021exploring}, RealForensics \cite{haliassos2022leveraging}, Lip Forensics \cite{haliassos2021lips}, EfficientNet+ViT \cite{coccomini2022combining}, Face X-Ray \cite{li2020face}, AltFreezing \cite{wang2023altfreezing}, CADMM \cite{dong2023implicit}, LAANet \cite{nguyen2024laa}, SBI \cite{shiohara2023blendface}, where source codes are available. We begin with a detailed overview of the deepfake videos used for evaluation, incorporating both state-of-the-art deepfake generation techniques with localized manipulations and traditional deepfake datasets. Then we provide an in-depth discussion of the experimental results, including self-analysis of the proposed method. We mainly highlight our model's superior detection capabilities on localized edits while maintaining strong generalization across diverse manipulations. Additional experiments are provided in the Appendix ~\ref{sec:APP_d}  .

\noindent\textbf{Latest Locally edited Deepfake methods:} We focus on locally edited deepfakes created using the latest state-of-the-art deepfake methods: Diffusion Video Autoencoders (DVA) \cite{kim2023diffusion}, Stitch It In Time (STIT) \cite{tzaban2022stitch}, and Disentangled Face Editing (DFE) \cite{yao2021latent}, Tokenflow \cite{TokenFlow}, VideoP2P \cite{Liu2023VideoP2PVE}, TextLive \cite{BarTal2022Text2LIVETL}, Fatezero \cite{Qi2023FateZeroFA}. DVA employs diffusion-based video editing to perform precise, text-guided alterations to the facial features in the diffusion feature space. In contrast, STIT and DFE exploit the rich latent space of StyleGAN for targeted facial modifications. Tokenflow, VideoP2P, FateZero and Text2Live adapts pre-trained text-to-image diffusion models for video editing while maintaining temporal coherence to allow text-guided edits. Refer to Figures ~\ref{fig:motexample} and \ref{fig:example} for examples of fake video frames that are visually indistinguishable from real ones, highlighting the superior  quality of these synthetic manipulations.   

To ensure comprehensive coverage of different facial manipulations, we incorporated a wide variety of facial features and attribute edits. For facial feature editing, we modified eye size, eye-eyebrow distance, nose ratio, nose-mouth distance, lip ratio, and cheek ratio. For facial attribute editing, we varied expressions such as smile, anger, disgust, and sadness. This diversity is essential for validating the robustness of our model over a wide range of localized edits. In total, we generated $50$ videos for each of the above-mentioned editing methods and validated our method's strong generalization for deepfake detection. Further details on the editing parameters for these deepfake generation methods are elaborated in Appendix ~\ref{APP_C}.

\noindent\textbf{Face swap and Face reenactment Methods:} 
For completeness, we also conduct evaluations on standard deepfake datasets, Celeb-DFv2 (CDF2) \cite{li2020celeb}, DeepFake Detection (DFD) \cite{google2024deepfake}, DeepFake Detection Challenge (DFDC) \cite{dolhansky2019deepfakedetectionchallengedfdc}, and WildDeepfake (DFW) \cite{zi2020wilddeepfake}. Note that these datasets consist of global  manipulations such as identity swaps and face reenactments. Additionally, the deepfake generation models used in the datasets are no longer state-of-the-art.

\noindent\textbf{Evaluation Metrics}
To evaluate our method against existing benchmarks and datasets, we use commonly applied standard evaluation metrics in deepfake detection: Area Under the Curve (AUC), which measures the discriminative capability of the model, and Average Precision (AP), which provides a precision-recall trade-off at the video level. More metric evaluations using Average Recall (AR) and mean F1-score (mF) are shown in Appendix ~\ref{sec:APP_d}.

\begin{table*}[h!]
\centering
\resizebox{\textwidth}{!}{%
\begin{tabular}{c|cc|cc|cc|cc|cc|cc|cc}
\hline
\textbf{Detection Methods} & \multicolumn{2}{c|}{\textbf{DVA}} & \multicolumn{2}{c|}{\textbf{STIT}} & \multicolumn{2}{c|}{\textbf{DFE}} & \multicolumn{2}{c|}{\textbf{Tokenflow}} & \multicolumn{2}{c|}{\textbf{VideoP2P}} & \multicolumn{2}{c|}{\textbf{TextLive}} & \multicolumn{2}{c}{\textbf{Fatezero}} \\
 & \multicolumn{2}{c|}{CVPR'23} & \multicolumn{2}{c|}{SIGGRAPH'22} & \multicolumn{2}{c|}{ICCV'21} & \multicolumn{2}{c|}{ICLR'23} & \multicolumn{2}{c|}{CVPR'24} & \multicolumn{2}{c|}{ECCV'22} & \multicolumn{2}{c}{ICCV'23} \\
\cline{2-15}
 & \textbf{AUC} & \textbf{AP} & \textbf{AUC} & \textbf{AP} & \textbf{AUC} & \textbf{AP} & \textbf{AUC} & \textbf{AP} & \textbf{AUC} & \textbf{AP} & \textbf{AUC} & \textbf{AP} & \textbf{AUC} & \textbf{AP} \\
\hline
FTCN & 27.1 & 30.1 & 34.8 & 37.0 & 33.5 & 35.8 & 29.5 & 32.1 & 31.0 & 33.7 & 30.2 & 32.9 & 28.7 & 31.3 \\
RealForensics & 37.5 & 40.3 & 46.9 & 49.8 & 45.6 & 48.6 & 41.2 & 44.2 & 42.8 & 45.8 & 42.0 & 45.1 & 40.5 & 43.4 \\
Lip Forensics & 33.8 & 37.1 & 42.0 & 45.8 & 40.9 & 44.6 & 36.5 & 40.3 & 38.0 & 41.9 & 37.3 & 41.2 & 35.8 & 39.5 \\
EfficientNet+ViT & 36.2 & 38.4 & 44.7 & 47.1 & 43.5 & 45.9 & 39.0 & 41.5 & 40.8 & 43.2 & 40.1 & 42.5 & 38.6 & 40.8 \\
Face X-Ray & 33.4 & 36.1 & 41.1 & 43.6 & 40.3 & 42.9 & 36.0 & 39.1 & 37.5 & 40.7 & 36.8 & 40.0 & 35.2 & 38.2 \\
LAA Net & 61.5 & 58.0 & 72.5 & 69.3 & 71.2 & 68.1 & 66.4 & 62.9 & 68.0 & 64.5 & 67.1 & 63.7 & 65.7 & 61.8 \\
SBI & \underline{65.2} & \underline{62.8} & \underline{75.5} & \underline{73.2} & \underline{73.3} & \underline{71.5} & \underline{69.0} & \underline{66.4} & \underline{70.8} & \underline{68.1} & \underline{70.2} & \underline{67.5} & \underline{68.6} & \underline{65.9} \\
AltFreezing & 41.1 & 44.0 & 51.8 & 53.6 & 51.2 & 52.8 & 45.6 & 47.1 & 47.5 & 48.9 & 46.9 & 48.3 & 45.0 & 46.5 \\
CADMM & 44.5 & 47.2 & 55.6 & 57.4 & 54.3 & 56.2 & 49.0 & 50.9 & 50.5 & 52.3 & 49.9 & 51.7 & 48.2 & 49.9 \\
\hline
\textbf{Ours} & \textbf{87.2} & \textbf{85.8} & \textbf{92.5} & \textbf{90.7} & \textbf{93.1} & \textbf{91.6} & \textbf{91.7} & \textbf{89.4} & \textbf{90.3} & \textbf{90.2} & \textbf{89.1} & \textbf{87.9} & \textbf{88.5} & \textbf{86.0} \\
\hline
\end{tabular}%
}
\caption{\textbf{Latest Locally edited Deepfake detection comparison:}: The proposed method demonstrates superior performance in detecting fake videos with latest deepfake generation methods, achieving at least $15 - 20\%$ increase in AUC and AP scores over the second-best method. Bold text indicates the best results, and underlined text indicates the second-best. }
\label{tab:localmanip}
\end{table*}

\begin{table}
\centering
\scalebox{0.6}{
\begin{tabular}{c|cc|cc|cc|cc}
\hline
Method & \multicolumn{2}{c|}{\textbf{CDF2}} & \multicolumn{2}{c|}{\textbf{DFD}} & \multicolumn{2}{c|}{\textbf{DFW}} & \multicolumn{2}{c}{\textbf{DFDC}} \\
\hline
 & \textbf{AUC} & \textbf{AP} & \textbf{AUC} & \textbf{AP} & \textbf{AUC} & \textbf{AP} & \textbf{AUC} & \textbf{AP} \\
\hline
FTCN \cite{zheng2021exploring} & $86.9$ & $86.0$ & $94.4$ & $90.33$ & $64.73$ & $65.5$ & $86.0$ & $87.48$  \\
RealForensics \cite{haliassos2022leveraging} & $85.6$ & $85.2$ & $82.24$ & $84.62$ & $66.72$ & $66.5$ & $89.7$ & $88.46$ \\
Lip Forensics \cite{haliassos2021lips} & $82.4$ & $82.66$ & $90.2$ & $89.37$ & $62.3$ & $60.75$ & $92.53$ & $93.41$  \\
EfficientNet+ViT \cite{coccomini2022combining} & $79.0$ & $75.61$ & $87.0$ & $88.09$ & $72.0$ & $68.74$ & $91.0$ & $85.12$  \\
Face X-Ray \cite{li2020face} & $79.5$ & $80.41$ & $95.4$ & $94.7$ & $61.5$ & $60.94$ & $85.27$ & $85.0$  \\
LAANet\cite{nguyen2024laa} & $\textbf{95.4}$ & $\mathbf{97.64}$ & $\mathbf{99.5}$ & $\mathbf{99.8}$ & $\underline{87.6}$ & $85.08$ & $86.94$ & $\mathbf{97.7}$   \\
SBI \cite{shiohara2023blendface} & $93.18$ & $85.16$ & $97.56$ & $92.79$ & $84.83$ & $\mathbf{88.37}$ & $86.16$ & $\underline{93.24}$ \\
AltFreezing \cite{wang2023altfreezing} & $89.5$ & $88.46$ & $\underline{98.50}$ & $93.17$ & $72.6$ & $72.0$ & $\mathbf{94.0}$ & $88.11$  \\
CADMM \cite{dong2023implicit} & $93.0$ & $91.12$ & $99.03$ & $\underline{99.59}$ & $75.0$ & $79.42$ & $88.3$ & $89.71$  \\
Ours & $\underline{93.84}$ & $\underline{95.27}$ & $97.15$ & $95.28$ & $\mathbf{91.0}$ & $\underline{88.25}$ & $\underline{93.0}$ & $91.93$  \\
\hline
\end{tabular}
}
\caption{\textbf{Traditional deepfake detection comparison:} AUC and AP scores are shown for traditional deepfake datasets. Our approach remains competitive with SOTA methods, underscoring its robustness and adaptability across diverse manipulations.}
\label{tab:cross_dataset_auc}
\end{table}

\subsection{Cross-Dataset Evaluation}
We first evaluate our model's generalization capability in a cross-dataset setting using latest deepfake generation methods \cite{kim2023diffusion,tzaban2022stitch,yao2021latent,TokenFlow,Liu2023VideoP2PVE,Qi2023FateZeroFA,BarTal2022Text2LIVETL} which involve local manipulations. As shown in Table \ref{tab:localmanip}, the existing SOTA detection methods, LAANet \cite{nguyen2024laa}, SBI \cite{shiohara2023blendface}, AltFreezing \cite{wang2023altfreezing} and CADMM \cite{dong2023implicit}, experience a significant drop in performance on the latest deepfake generation methods. The current SOTA methods exhibit AUCs as low as $48\mbox{-}71\%$, demonstrating their poor generalization capabilities to the recent deepfakes. On the other hand, our method demonstrates robust generalization, achieving an AUC in the range $87$-$ 93\%$. A similar trend is noticeable in the case of average precision as well. As shown in Table~\ref{tab:cross_dataset_auc}, our method also consistently achieves high performance on standard datasets, exceeding $90\%$ AUC and are competitive with recent deepfake detection models. We highlight that these standard datasets primarily contain deepfake methods introduced before 2020, prior to the emergence of recent video editing techniques for deepfake generation. 

To visually illustrate our model's superior performance, we display frames of  videos with various localized edits in Fig.~\ref{fig:example}, along with probability scores for detection. Our method consistently achieves confidence scores exceeding $90\%$ in detecting localized edits within fake videos, whereas existing state-of-the-art detection methods fall below $50\%$, underscoring the robustness of our approach. The demonstrated results highlight the limitations of current deepfake detection techniques in handling localized edits, the strong sensitivity of our model to fine-grained manipulations, and the generalizable property of the proposed method. 


\begin{table}[!htp]
\centering
\resizebox{\columnwidth}{!}{ 
\begin{tabular}{l|ccccccc}
\hline
\textbf{Perturbation} & \textbf{DVA} & \textbf{STIT} & \textbf{DFE} & \textbf{TF} & \textbf{T2L} & \textbf{FZ} & \textbf{V2P} \\
\hline
Gaussian Noise & 82.3 & 83.1 & 83.8 & 84.5 & 85.0 & 84.2 & 83.7 \\
Saturation Change & 87.2 & 88.1 & 88.5 & 88.9 & 89.3 & 89.0 & 88.6 \\
Blockwise Distortion & 86.7 & 87.5 & 87.8 & 88.3 & 88.6 & 88.4 & 88.0 \\
Contrast Change & 86.8 & 87.6 & 88.0 & 88.4 & 88.8 & 88.5 & 88.1 \\
Pixelation & 86.0 & 86.8 & 87.3 & 87.8 & 88.1 & 87.9 & 87.5\\
Gaussian Blur & 86.3 & 87.0 & 87.5 & 87.9 & 88.2 & 88.0 & 87.6\\
No Perturbation & \textbf{88.5} & \textbf{89.3} & \textbf{89.7} & \textbf{90.1} & \textbf{90.6} & \textbf{90.3} & \textbf{89.8}\\
\hline
\end{tabular}
}
\caption{\textbf{Effect of perturbations on AUC:} The proposed method shows resilience to various perturbations, with only a slight reduction in AUC. The highest reduction in detection performance is observed when Gaussian noise is added to videos.}
\label{tab:perturbations}
\end{table}
\subsection{Robustness to Perturbrations} 
In general, adversarial attacks involve introducing different perturbations such as noise, blur, etc., or changing video properties such as saturation, contrast, etc. A generalizable deepfake detection algorithm should be robust against such alterations. Similar to \cite{dong2023implicit}, we applied the following standard perturbations: 1) Saturation in a range of $0.5$ to $2.0$, 2) Contrast in a range of $0.5$ to $2.0$, 3) Gaussian Noise with a standard deviation of $0.01$ to $0.1$, 4) Blur with a Gaussian kernel size of radius $3$ to $11$, 5) Pixelation with downsampling factors  $4$ to $16$, and 6) Blocking artifacts with quality levels from $10$ to $50$. In Table~\ref{tab:perturbations}, the AUC scores of our model are  evaluated on deepfake generation methods (averaged over $50$ videos), after applying these perturbations to the test videos. Results show that Gaussian noise resulted in a slight decrease in AUC scores, while other alterations had minimal impact on model performance, demonstrating the robustness of our method against various distortions.

\begin{figure}[!h]
    \centering
    \begin{subfigure}[c]{0.09\textwidth}  
        \centering
        \includegraphics[width=\textwidth]{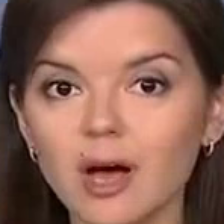}        
    \end{subfigure}
    \begin{subfigure}[c]{0.09\textwidth}  
        \centering
        \includegraphics[width=\textwidth]{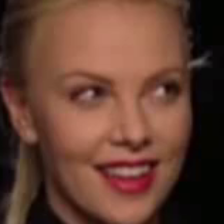} 
    \end{subfigure}
    \begin{subfigure}[c]{0.09\textwidth}  
        \centering
        \includegraphics[width=\textwidth]{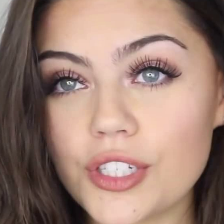} 
    \end{subfigure}
    \begin{subfigure}[c]{0.09\textwidth}  
        \centering
        \includegraphics[width=\textwidth]{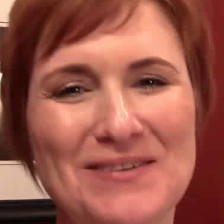} 
    \end{subfigure}
    \begin{subfigure}[c]{0.09\textwidth}  
        \centering
        \includegraphics[width=\textwidth]{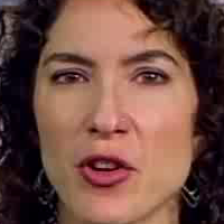} 
    \end{subfigure}
    
    \vspace{1mm}
    \begin{subfigure}[c]{0.09\textwidth}  
        \centering
        \includegraphics[width=\textwidth]{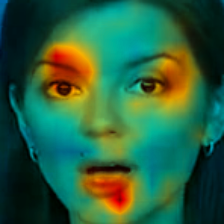}   
        \caption*{Shock}
    \end{subfigure}
    \begin{subfigure}[c]{0.09\textwidth}  
        \centering
        \includegraphics[width=\textwidth]{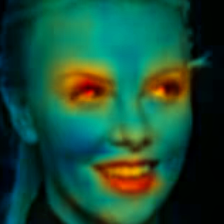} 
        \caption*{Gaze}
    \end{subfigure}
    \begin{subfigure}[c]{0.09\textwidth}  
        \centering
        \includegraphics[width=\textwidth]{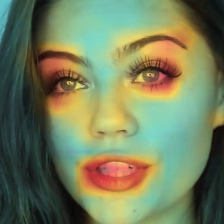} 
        \caption*{Eye Raise}
    \end{subfigure}
    \begin{subfigure}[c]{0.09\textwidth}  
        \centering
        \includegraphics[width=\textwidth]{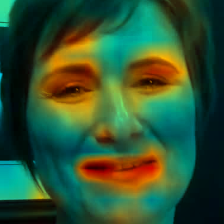} 
        \caption*{Smile}
    \end{subfigure}
    \begin{subfigure}[c]{0.09\textwidth}  
        \centering
        \includegraphics[width=\textwidth]{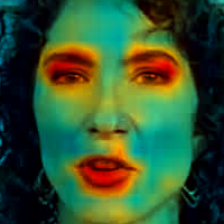} 
        \caption*{Angry}
    \end{subfigure}        
    \caption{\textbf{Local Attention Visualization:} Overlay of attention maps (bottom row) from the final cross-attention block in AUGFE on facial video frames (top row). The maps highlight key action units, demonstrating our model's ability to consistently capture critical facial features across diverse expressions.}
\label{fig:attention_map}
\end{figure}

\subsection{Ablation Study}

We first perform a detailed interpretability analysis to assess whether our modified encoder effectively focuses on key action units within face-centered videos. In Fig.~\ref{fig:attention_map}, the attention maps from the final cross-attention block of our fused encoder highlight the model's ability to capture local dynamics, using multiple action units. This analysis visually substantiates why the latent output representation of our constructed fused encoder are robust enough to detect deepfake videos with localized manipulations.

We next conduct an in-depth analysis by systematically selecting different action units, as presented in Table~\ref{tab:au_ablation}, demonstrating that our chosen AUs yield the best detection performance. We use the open-source benchmark ~\cite{9577264} to generate ground truth AU maps for the first 16 AUs, which effectively capture key facial feature information. This evaluation aligns with the attention maps illustrated in Fig.~\ref{fig:attention_map}, where our method effectively captures most of the important facial expressions by leveraging comprehensive information across action units. 

In Table \ref{tab:ablation_e2_new}, we highlight the need for a cross-attention based fusion of latent representations from two pretext tasks for deepfake detection. Note that our model fuses latent embeddings from two encoders, VFE pretrained by masking, and AUE pretrained by AU detection. To assess the performance of each encoder independently and understand the impact of fusing VFE layer features with AUE layer features  through cross-attention, we conduct  experiments where each encoder is fine-tuned separately for the downstream task of deepfake detection. Note that the baseline here is a model analogous to the architecture of VFE without any pre-training.

To highlight the contribution of the AU encoder (AUE) guidance during fine-tuning, we first remove the connections from AUE and finetune only the frame encoder (VFE)  for deepfake detection. As shown in Table~\ref{tab:ablation_e2_new}, this leads to an  unacceptable drop in performance across all the state-of-the-art deepfake generation modes, since these method focus on localized edit patterns. These findings underscore the critical role of AU guidance from AUE in detecting subtle manipulations. Next, to assess the importance of VFE, we  fine-tune only the AU encoder (AUE) for deepfake detection. As shown in Table~\ref{tab:ablation_e2_new}, though the model with only AUE performs moderately well, it underperforms considerably compared to the full model with both encoders. This is likely due to the absence of VFE, which captures global facial representations that are crucial for effectively distinguishing between real and fake videos. These results suggest that the cross-attention mechanism between VFE and AUE is crucial for detecting local facial manipulations. More experiments on self-analysis of the proposed method is provided in Appendix~\ref{APP_B}.

\begin{table}
\centering
\setlength{\tabcolsep}{2pt} 
\scriptsize 
\resizebox{\linewidth}{!}{%
\begin{tabular}{l|ccccccc}
\toprule
\textbf{AU Set} & \textbf{DVA} & \textbf{STIT} & \textbf{DFE} & \textbf{TF} & \textbf{FZ} & \textbf{VP2P} & \textbf{T2L} \\
\midrule
AU's 1-5 Eyes   & 63.1  & 72.4  & 72.8  & 71.4  & 64.3  & 67.7  & 66.2 \\
AU's 7-11 Nose  & 66.1  & 73.1  & 73.8  & 72.4  & 67.3  & 70.7  & 69.2  \\
AU's 12-16 Lips & 72.4  & 79.3  & 80.0  & 78.6  & 73.1  & 76.9  & 75.4  \\
AU's 1-16 (All) & 87.20 & 92.50 & 93.10 & 91.70 & 88.50 & 90.30 & 89.10 \\
\bottomrule
\end{tabular}%
}
\caption{\textbf{Effect of Different AU Subsets}: The selected set of action units is essential for effectively capturing comprehensive facial dynamics, thereby enhancing the deepfake detection accuracy.}
\label{tab:au_ablation}
\end{table}

\begin{table}
\centering
\resizebox{\columnwidth}{!}{%
\begin{tabular}{c|ccccccc}
\hline
\textbf{Components} & \textbf{DVA} & \textbf{STIT} & \textbf{DFE} & \textbf{TF} & \textbf{FZ} & \textbf{VP2P} & \textbf{TL} \\
\hline
W/O VFE and AUE & $35.10$ & $32.10$ & $33.80$ & $38.0$ & $36.30$ & $39.70$ & $38.0$ \\
\hline
With VFE & $52.60$ & $62.60$ & $65.30$ & $60.90$ & $54.80$ & $57.70$ & $56.0$ \\
\hline
With AUE & $75.10$ & $82.10$ & $82.80$ & $81.80$ & $76.30$ & $79.50$ & $78.20$ \\
\hline
Fused encoder & $\mathbf{87.20}$ & $\mathbf{93.50}$ & $\mathbf{93.10}$ & $\mathbf{91.30}$ & $\mathbf{88.50}$ & $\mathbf{91.70}$ & $\mathbf{90.1}$ \\
\hline
\end{tabular}
}
\caption{\textbf{Effect of Encoders:} Performance degradation without VFE and AUE shows their importance. Our approach of fusing two encoders achieves the best performance.}
\label{tab:ablation_e2_new}
\end{table}

\section{Conclusion}
In this work, we present a novel deepfake detection method specifically designed to identify subtle, localized manipulations in video, addressing a challenge overlooked by existing methods.  To the best of our knowledge, this is the first approach that effectively targets localized edits, leveraging a powerful fusion of spatio-temporal representations from two complementary pretext tasks-masked frame reconstruction and facial action unit detection. Our method significantly outperforms existing state-of-the-art deepfake detection models in identifying latest deepfakes localized manipulations. Additionally, it demonstrates competitive performance with top methods on standard deepfake datasets involving global alterations. These results validate our approach as highly effective across diverse deepfake types and hence robustly generalizable. Future work will explore extending our fused latent video representation framework to additional downstream tasks in video analysis, expanding beyond face-centered content.

\small
\bibliographystyle{ieeenat_fullname}
\bibliography{refs}

\clearpage
\appendix

\section*{Appendix}
In this section, we provide additional details to demonstrate the  effectiveness of the proposed method. We start by providing detailed training procedure for both the pretext tasks and the deepfake detection framework.  Further, we evaluate each pretext task individually through reconstruction performance and visual comparisons. We also include additional metrics, such as Average Recall (AR) and mean F1 score (mF1), along with the primary metrics of AUC and Average Precision (AP) shown in the manuscript. This offers a more detailed comparison of our method against existing state-of-the-art detection methods across different deepfake generators. In addition, we elaborate on the construction of the latest deepfake videos with local manipulations, including descriptions of the methods and parameters used to generate fake videos with localized subtle edits.

\section{Comprehensive Training Details}
\label{APP_a}
The two pretext tasks are trained using the CelebV-HQ dataset \cite{zhu2022celebv}, which contains approximately $35,000$ real facial videos. The first pretext model for the reconstruction of masked frames minimizes a variant of $\ell_1$ reconstruction loss (Huber loss) between ground truth frames and frames reconstructed from masked inputs, following a VideoMAE-like approach \cite{tong2022videomae}. The second pretext model for Facial Action Unit (AU) detections is trained using Huber loss between predicted AU maps and ground truth maps, to predict $16$ AUs for each frame. To generate the ground-truth attention map for every action unit, we define landmarks corresponding to the different AUs  following the conventional approach in \cite{shao2018deep, li2018eac}. Elliptical regions are fitted to these landmarks as initial AU regions, which are then smoothed using a Gaussian filter of radius $3$. This process yields $16$ distinct AU maps, each corresponding to a specific localized action, for a single frame.
\begin{table}[h]
\centering
\footnotesize
\resizebox{\columnwidth}{!}{%
\begin{tabular}{|c|c|c|c|c|c|c|c|}
\hline
\textbf{} & \textbf{DVA} & \textbf{STIT} & \textbf{DFE} & \textbf{TF} & \textbf{FZ} & \textbf{T2L} & \textbf{V2P} \\
\hline
MAE: Random Masking  & 2.71e\text{-}8 & 2.6e\text{-}8 & 2.64e\text{-}8 & 2.48e\text{-}8 & 2.35e\text{-}8 & 2.40e\text{-}8 & 2.38e\text{-}8 \\
\hline
MAE: AU Detection  & 1.23e\text{-}8 & 1.17e\text{-}8 & 1.19e\text{-}8 & 1.05e\text{-}8 & 9.8e\text{-}9 & 1.02e\text{-}8 & 1.01e\text{-}8 \\
\hline
\end{tabular}%
}
\caption{\textbf{Reconstruction Error for Pretext Tasks:} The pretext models for random masking and AU detection are evaluated independently, with MAE between ground truth and reconstructions tabulated across datasets (normalized between $0$ and $1$). The negligible MAE confirms their effectiveness in both tasks.}
\label{tab:reconstrerror}
\end{table}

We trained both the pretext models using the Adam optimizer with a batch size of $8$ for $600$ epochs. Gradient accumulation was applied every $20$ steps. We used the pre-trained checkpoints from VideoMAE \cite{tong2022videomae} to initialize our weights for both the pretext tasks.

During fine-tuning for deepfake detection, the fused encoder shown in Fig.~$4$ in the manuscript, is trained with a classifier on the FF++ dataset \cite{rossler2019faceforensics}, consisting of $700$ real and $3,600$ fake videos generated via four manipulation methods \cite{kowalski2024faceswap,thies2016face2face,deepfakes_github,thies2019deferred}. Focal Loss \cite{ross2017focal} is used to address class imbalance. For finetuning, we used a batch size of $8$ for $100$ epochs. A learning rate of $1e\mbox{-}5$ with an  exponential decay of $1e\mbox{-}3$ is used for both the pretext tasks and the  finetuning stage.



\begin{table*}[h!]
\centering
\resizebox{\textwidth}{!}{%
\begin{tabular}{c|cccc|cccc|cccc|cccc|cccc|cccc|cccc}
\hline
\textbf{Detection Methods} & \multicolumn{4}{c|}{\textbf{DVA}} & \multicolumn{4}{c|}{\textbf{STIT}} & \multicolumn{4}{c|}{\textbf{DFE}} & \multicolumn{4}{c|}{\textbf{Tokenflow}} & \multicolumn{4}{c|}{\textbf{VideoP2P}} & \multicolumn{4}{c|}{\textbf{TextLive}} & \multicolumn{4}{c}{\textbf{Fatezero}} \\
 & \textbf{AUC} & \textbf{AP} & \textbf{AR} & \textbf{mf} & \textbf{AUC} & \textbf{AP} & \textbf{AR} & \textbf{mf} & \textbf{AUC} & \textbf{AP} & \textbf{AR} & \textbf{mf} & \textbf{AUC} & \textbf{AP} & \textbf{AR} & \textbf{mf} & \textbf{AUC} & \textbf{AP} & \textbf{AR} & \textbf{mf} & \textbf{AUC} & \textbf{AP} & \textbf{AR} & \textbf{mf} & \textbf{AUC} & \textbf{AP} & \textbf{AR} & \textbf{mf} \\
\hline
FTCN & 27.1 & 30.1 & 25.4 & 27.2 & 34.8 & 37.0 & 32.1 & 34.5 & 33.5 & 35.8 & 30.9 & 33.3 & 29.5 & 32.1 & 27.0 & 29.5 & 31.0 & 33.7 & 29.2 & 31.5 & 30.2 & 32.9 & 28.6 & 30.8 & 28.7 & 31.3 & 26.9 & 29.1 \\
RealForensics & 37.5 & 40.3 & 36.2 & 38.2 & 46.9 & 49.8 & 44.5 & 47.1 & 45.6 & 48.6 & 43.2 & 45.8 & 41.2 & 44.2 & 39.0 & 41.6 & 42.8 & 45.8 & 40.4 & 43.0 & 42.0 & 45.1 & 39.8 & 42.3 & 40.5 & 43.4 & 38.5 & 40.9 \\
Lip Forensics & 33.8 & 37.1 & 32.4 & 34.7 & 42.0 & 45.8 & 40.3 & 43.0 & 40.9 & 44.6 & 39.0 & 41.7 & 36.5 & 40.3 & 34.7 & 37.4 & 38.0 & 41.9 & 36.0 & 38.8 & 37.3 & 41.2 & 35.4 & 38.2 & 35.8 & 39.5 & 33.9 & 36.6 \\
EfficientNet+ViT & 36.2 & 38.4 & 34.9 & 36.6 & 44.7 & 47.1 & 42.8 & 44.9 & 43.5 & 45.9 & 41.3 & 43.5 & 39.0 & 41.5 & 37.2 & 39.3 & 40.8 & 43.2 & 38.6 & 40.8 & 40.1 & 42.5 & 38.0 & 40.2 & 38.6 & 40.8 & 36.5 & 38.6 \\
Face X-Ray & 33.4 & 36.1 & 31.5 & 33.8 & 41.1 & 43.6 & 38.8 & 41.2 & 40.3 & 42.9 & 38.2 & 40.5 & 36.0 & 39.1 & 34.2 & 36.5 & 37.5 & 40.7 & 35.4 & 37.8 & 36.8 & 40.0 & 34.8 & 37.3 & 35.2 & 38.2 & 33.5 & 35.8 \\
LAA Net & 61.5 & 58.0 & 55.2 & 56.6 & 72.5 & 69.3 & 66.4 & 67.8 & 71.2 & 68.1 & 64.8 & 66.3 & 66.4 & 62.9 & 60.3 & 61.6 & 68.0 & 64.5 & 62.0 & 63.3 & 67.1 & 63.7 & 60.9 & 62.2 & 65.7 & 61.8 & 59.3 & 60.6 \\
SBI & \underline{65.2} & \underline{62.8} & \underline{59.4} & \underline{61.0} & \underline{75.5} & \underline{73.2} & \underline{70.1} & \underline{71.6} & \underline{73.3} & \underline{71.5} & \underline{68.5} & \underline{69.9} & \underline{69.0} & \underline{66.4} & \underline{63.5} & \underline{64.9} & \underline{70.8} & \underline{68.1} & \underline{65.2} & \underline{66.6} & \underline{70.2} & \underline{67.5} & \underline{64.7} & \underline{66.1} & \underline{68.6} & \underline{65.9} & \underline{63.1} & \underline{64.5} \\
\textbf{Ours} & \textbf{87.2} & \textbf{85.8} & \textbf{82.5} & \textbf{84.1} & \textbf{92.5} & \textbf{90.7} & \textbf{88.1} & \textbf{89.4} & \textbf{93.1} & \textbf{91.6} & \textbf{89.5} & \textbf{90.5} & \textbf{91.7} & \textbf{89.4} & \textbf{87.0} & \textbf{88.2} & \textbf{90.3} & \textbf{90.2} & \textbf{86.9} & \textbf{88.0} & \textbf{89.1} & \textbf{87.9} & \textbf{85.5} & \textbf{86.6} & \textbf{88.5} & \textbf{86.0} & \textbf{84.2} & \textbf{85.1} \\
\hline
\end{tabular}%
}
\caption{\textbf{Cross-Dataset Quantitative Comparison:} AUC, AP, AR, and mF1 scores evaluated across the latest deepfake generation methods. The results highlight the superior detection performance of our method, significantly surpassing existing state-of-the-art approaches in identifying fine-grained localized edits.}
\label{tab:cross_dataset_ours}
\end{table*}

\begin{table*}
\centering
\resizebox{\textwidth}{!}{%
\begin{tabular}{|c|cccc|cccc|cccc|cccc|}
\hline
 \textbf{Method}  & \multicolumn{4}{c|}{\textbf{CDF2}} & \multicolumn{4}{c|}{\textbf{DFD}} & \multicolumn{4}{c|}{\textbf{DFW}} & \multicolumn{4}{c|}{\textbf{DFDC}} \\
\hline
 & \textbf{AUC} & \textbf{AP} & \textbf{AR} & \textbf{mf} & \textbf{AUC} & \textbf{AP} & \textbf{AR} & \textbf{mf} & \textbf{AUC} & \textbf{AP} & \textbf{AR} & \textbf{mf} & \textbf{AUC} & \textbf{AP} & \textbf{AR} & \textbf{mf}  \\
\hline
LAA Net \cite{nguyen2024laa} & $\textbf{95.4}$ & $\mathbf{97.64}$ & $\underline{87.71}$ & $\textbf{92.41}$ & $\mathbf{99.5}$ & $\mathbf{99.8}$ & $95.47$ & $\textbf{97.59}$ & $\underline{87.6}$ & $85.08$ & $69.66$ & $78.56$ & $86.94$ & $\mathbf{97.7}$ & $73.37$ & $83.81$ \\
SBI \cite{shiohara2023blendface} & $93.18$ & $85.16$ & $82.68$ & $83.90$ & $97.56$ & $92.79$ & $89.49$ & $91.11$ & $84.83$ & $\textbf{88.37}$ & $\underline{81.64}$ & $\underline{84.60}$ & $86.16$ & $\underline{93.24}$ & $71.58$ & $80.99$ \\
AltFreezing \cite{wang2023altfreezing} & $89.5$ & $88.46$ & $85.50$ & $86.24$ & $98.50$ & $97.86$ & $\underline{97.0}$ & $\underline{97.41}$ & $72.6$ & $70.86$ & $68.5$ & $69.66$ & $\textbf{94.0}$ & $92.57$ & $\textbf{91.1}$ & $\textbf{91.80}$ \\
CADMM \cite{dong2023implicit} & $93.0$ & $91.12$ & $77.00$ & $83.46$ & $99.03$ & $\underline{99.59}$ & $82.17$ & $90.04$ & $75.0$ & $72.80$ & $71.26$ & $72.14$ & $88.3$ & $86.7$ & $85.62$ & $86.1$ \\
EfficientNet+ViT \cite{coccomini2022combining} & $79.0$ & $75.61$ & $74.5$ & $75.05$ & $87.0$ & $88.09$ & $85.8$ & $86.93$ & $72.0$ & $68.74$ & $67.0$ & $67.85$ & $91.0$ & $85.12$ & $83.7$ & $84.39$ \\
Ours & $\underline{93.84}$ & $\underline{95.27}$ & $\textbf{92.66}$ & $\underline{92.17}$ & $97.15$ & $95.28$ & $\textbf{98.6}$ & $97.23$ & $\mathbf{91.0}$ & $\underline{88.25}$ & $\textbf{88.63}$ & $\textbf{87.40}$ & $\underline{93.0}$ & $91.93$ & $\underline{90.38}$ & $\underline{91.26}$ \\
\hline
\end{tabular}
}
\caption{\textbf{Cross-Dataset Quantitative Comparison:} Evaluation of AUC, AP, AR, and mF1 scores across standard deepfake datasets, focused on face swapping and reenactment. The proposed method is competitive with existing SOTA approaches across all metrics. Notably, our method achieves superior AR values, indicating high sensitivity in detecting fake videos (positives).}
\label{tab:cross_dataset_auc_extended_filled}
\end{table*}

\vspace{-1.8 mm}
\section{Evaluation on Pretext tasks}
\label{APP_B}
We evaluate the performance of pretext models independently to demonstrate the effectiveness of the representations learned by the respective encoders, VFE (Video Frame Encoder) and AUE (Action Unit Encoder). For the first self-supervised task -  reconstruction of face-centered frames from masked input frames - we compute the Mean Absolute Error (MAE) between the output reconstructed frames and the ground truth. MAE is first computed across all 16 input frames for each video to obtain a video-level MAE. This score is then averaged over all videos across diverse methods, and presented in the first row of Table~\ref{tab:reconstrerror}.

For the second self-supervised pretext task - reconstruction of AU maps for each video - we compute the MAE between the $16$ reconstructed AU maps and the corresponding ground truth maps for every frame. For diverse methods, this per-frame MAE is initially averaged across all 16 frames for each video. These video-level MAE values are then averaged across the all the videos corresponding to a particular deepfake generation method to obtain the final reconstruction error, as reported in the second row of Table~\ref{tab:reconstrerror}. The low MAE values for both the pretext tasks demonstrates effectiveness of their respective learned representations. In Fig.~\ref{fig:audetection}, a qualitative comparison is shown, where, for a single frame, we display selected ground-truth AU maps alongside their reconstructed counterparts as output by the model. These visualizations highlight the model's capability in accurately capturing fine-grained facial details.

\begin{table}
    \centering
    \begin{tabular}{>{\centering\arraybackslash}m{0.09\textwidth} @{\hskip 3pt} 
                    >{\centering\arraybackslash}m{0.09\textwidth} @{\hskip 3pt} 
                    >{\centering\arraybackslash}m{0.09\textwidth} @{\hskip 3pt} 
                    >{\centering\arraybackslash}m{0.09\textwidth} @{\hskip 3pt}
                    >{\centering\arraybackslash}m{0.09\textwidth}}
        \includegraphics[width=0.95\linewidth]{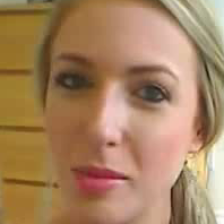} & 
        \includegraphics[width=0.95\linewidth]{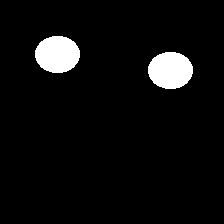} & 
        \includegraphics[width=0.95\linewidth]{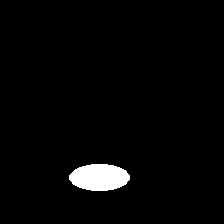} & 
        \includegraphics[width=0.95\linewidth]{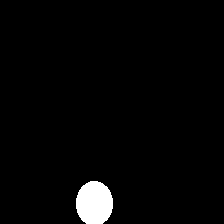} & 
        \includegraphics[width=0.95\linewidth]{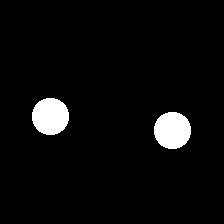} \\ 
        & \includegraphics[width=0.95\linewidth]{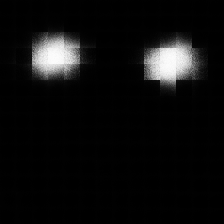} & 
        \includegraphics[width=0.95\linewidth]{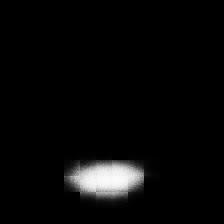} & 
        \includegraphics[width=0.95\linewidth]{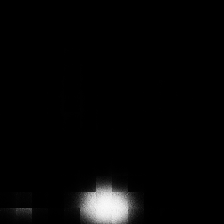} & 
        \includegraphics[width=\linewidth]{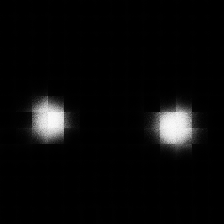} \\ 
    \end{tabular}
    \captionof{figure}{\textbf{AU Detection Maps Comparison:} Comparison of four AU maps for a sample test image, with ground-truth maps (top row) and reconstructed maps (bottom row). The accurate reconstruction across action units highlights the effectiveness of the pretext task in preserving spatio-temporal localization.}
    \label{fig:audetection}
\end{table}

\section{Latest Locally edited Deepfake Videos}
\label{APP_C}
We leveraged seven state-of-the-art methods to test proposed deepfake detection method : Diffusion Video Autoencoders (DVA) \cite{kim2023diffusion}, Stitch It In Time (STIT)  \cite{tzaban2022stitch}, Disentangled Face Editing (DFE) \cite{yao2021latent}, Tokenflow \cite{TokenFlow}, VideoP2P \cite{Liu2023VideoP2PVE}, FateZero \cite{Qi2023FateZeroFA}, Text2Live \cite{BarTal2022Text2LIVETL}. For all the methods, we utilized their official source code and generated $50$ videos each. These methods enabled localized edits targeting eyes, mouth, expressions, age, and gender transformations. For DVA, we used $1000$ sampling steps, a learning rate of $0.002$ (for finetuning), and an editing scale of $0.5$. For StyleGAN2~\cite{karras2020analyzing} based editing methods STIT and DFE, we followed the common pipeline for editing, which involves video inversion to latent space, finetuning the generator for a specific video, and editing the latent vector. For STIT, we used $50$ steps for finetuning the generator, along with an editing range of $+6$ to $-6$. Similarly, for DFE, we used $50$ steps for finetuning and edtiting range between $-10$ to $+10$. TokenFlow, Video-P2P, and FateZero utilize pre-trained diffusion models during inference, standardized with 50 DDIM inversion steps and a classifier-free guidance scale of 7.5 for text fidelity. Video-P2P further employs a cross-attention replacement ratio of 0.4 to enhance temporal consistency.
Text2LIVE, in contrast is trained for each video using a video-specific generator for 1,000 optimization steps.

\begin{table}
    \centering
    \begin{tabular}{>{\centering\arraybackslash}m{0.15\textwidth} @{\hskip 3pt} 
                    >{\centering\arraybackslash}m{0.15\textwidth} @{\hskip 3pt}
                    >{\centering\arraybackslash}m{0.15\textwidth}}
        \includegraphics[width=0.95\linewidth]{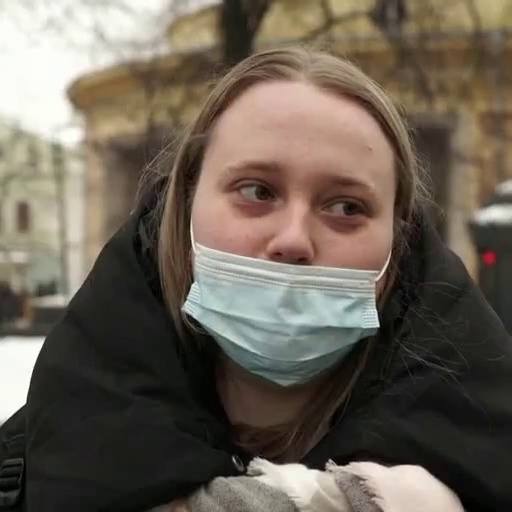} & 
        \includegraphics[width=0.95\linewidth]{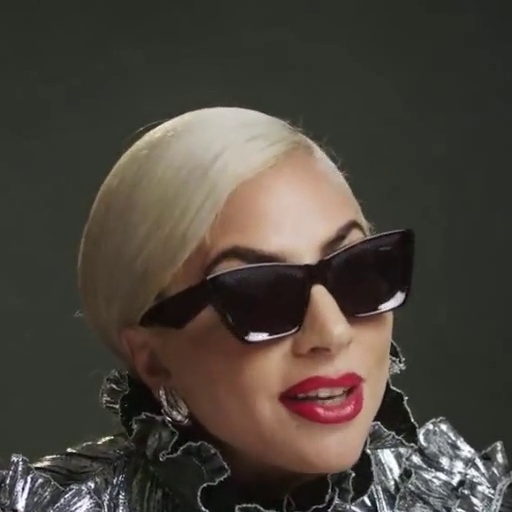} & 
        \includegraphics[width=0.95\linewidth]{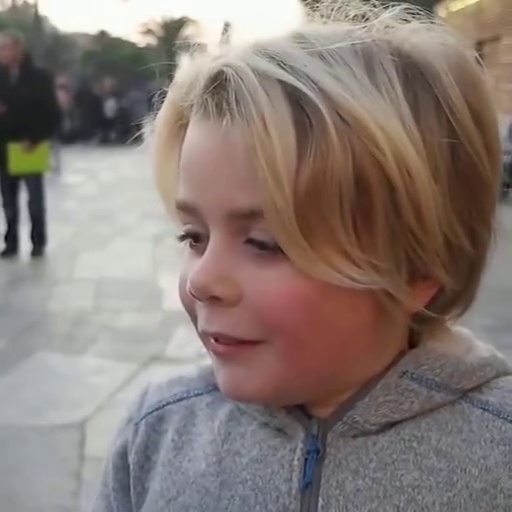} \\ 
        \includegraphics[width=0.95\linewidth]{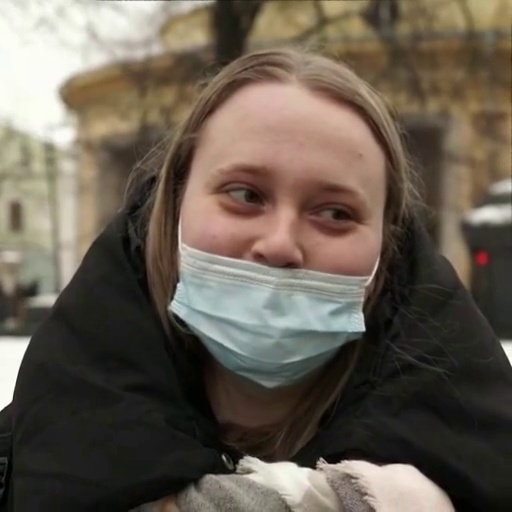} & 
        \includegraphics[width=0.95\linewidth]{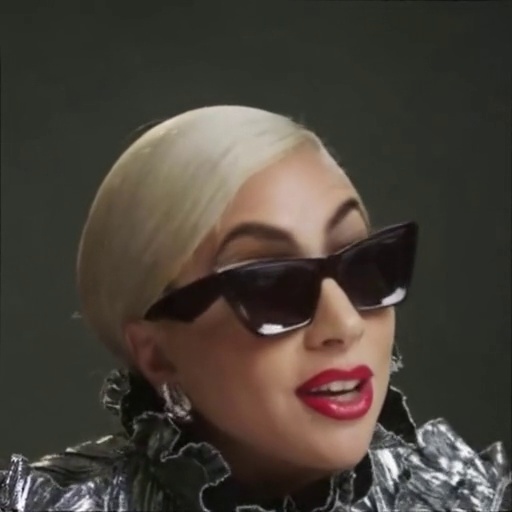} & 
        \includegraphics[width=0.95\linewidth]{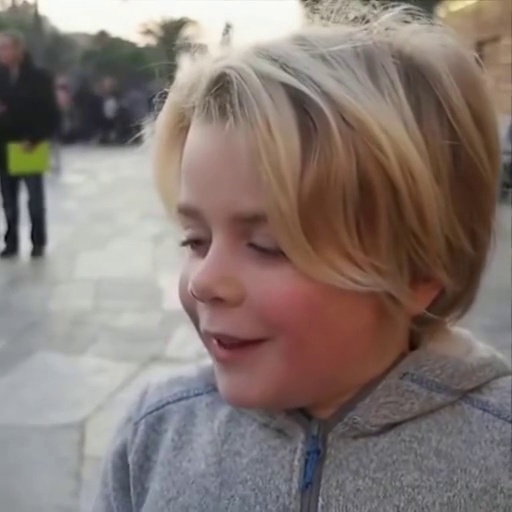} \\ 
        \small \textbf{Fake Score: $56.6$} & \small \textbf{Fake Score: $60.2$} & \small \textbf{Fake Score: $67.3$}
    \end{tabular}
    \captionof{figure}{\textbf{Limitations:} Detection of fake videos (bottom row) generated from real videos (top row) with localized edits. A noticeable drop in confidence scores ($20\mbox{-}35\%$) is observed in case of occlusions or side-facing poses, since the proposed representations do not capture action unit dynamics effectively.
    }
    \label{fig:failure_cases}
\end{table}

\begin{table*}
    \centering
    \begin{tabular}{>{\centering\arraybackslash}m{0.18\textwidth} @{\hskip 3pt} 
                     @{\hskip 3pt} >{\centering\arraybackslash}m{0.18\textwidth} 
                    @{\hskip 0.5pt} >{\centering\arraybackslash}m{0.18\textwidth} 
                    @{\hskip 0.5pt} >{\centering\arraybackslash}m{0.18\textwidth} @{\hskip 3pt}
                     @{\hskip 3pt} >{\centering\arraybackslash}m{0.22\textwidth}}
        \includegraphics[width=0.95\linewidth]{Figures/Video8/img_8_original.png} & 
        \includegraphics[width=0.95\linewidth]{Figures/Video8/img_8_gender.png} & 
        \includegraphics[width=0.95\linewidth]{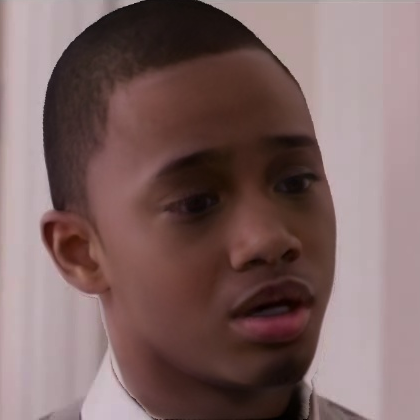} & 
        \includegraphics[width=0.95\linewidth]{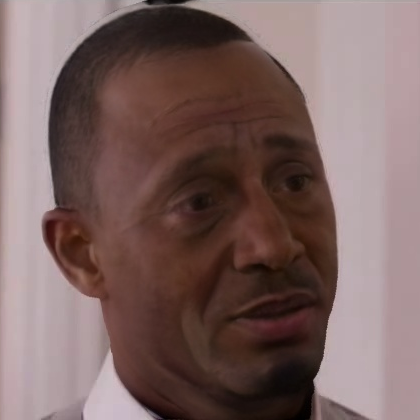} & 
        \textit{\textbf{\footnotesize Gender, Young, Old}} 
        \includegraphics[width=\linewidth]{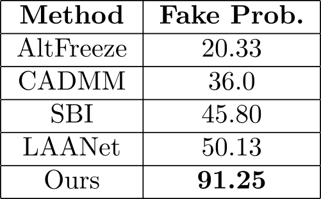} \\ 
        \includegraphics[width=0.95\linewidth]{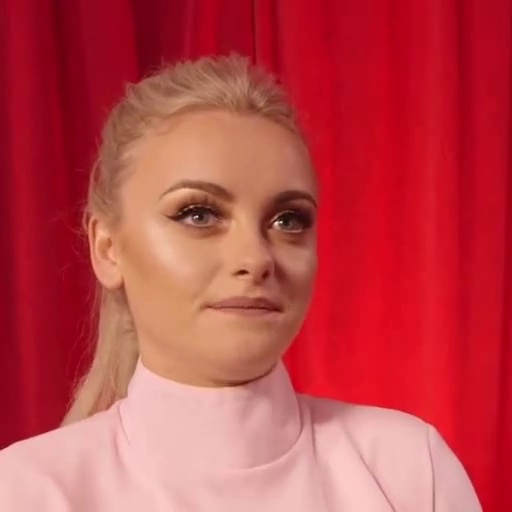} & 
        \includegraphics[width=0.95\linewidth]{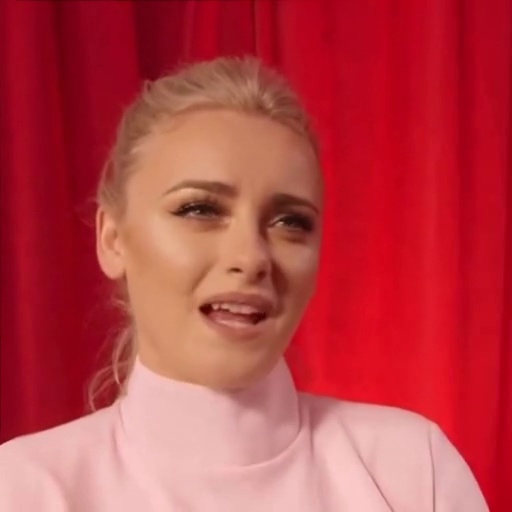} & 
        \includegraphics[width=0.95\linewidth]{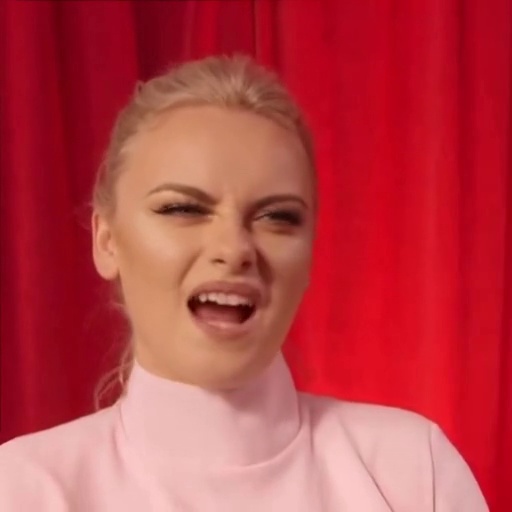} & 
        \includegraphics[width=0.95\linewidth]{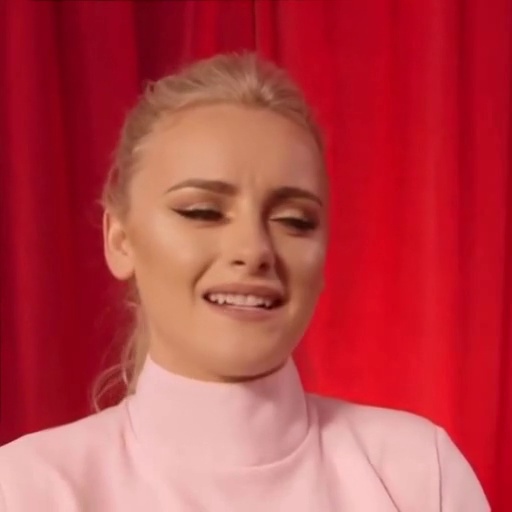} & 
        \textit{\textbf{\footnotesize Sad, Anger, Disgust}} 
        \includegraphics[width=\linewidth]{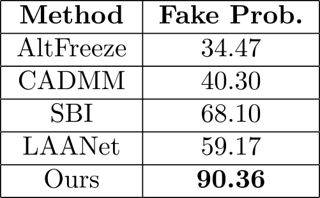} \\ 
        \includegraphics[width=0.95\linewidth]{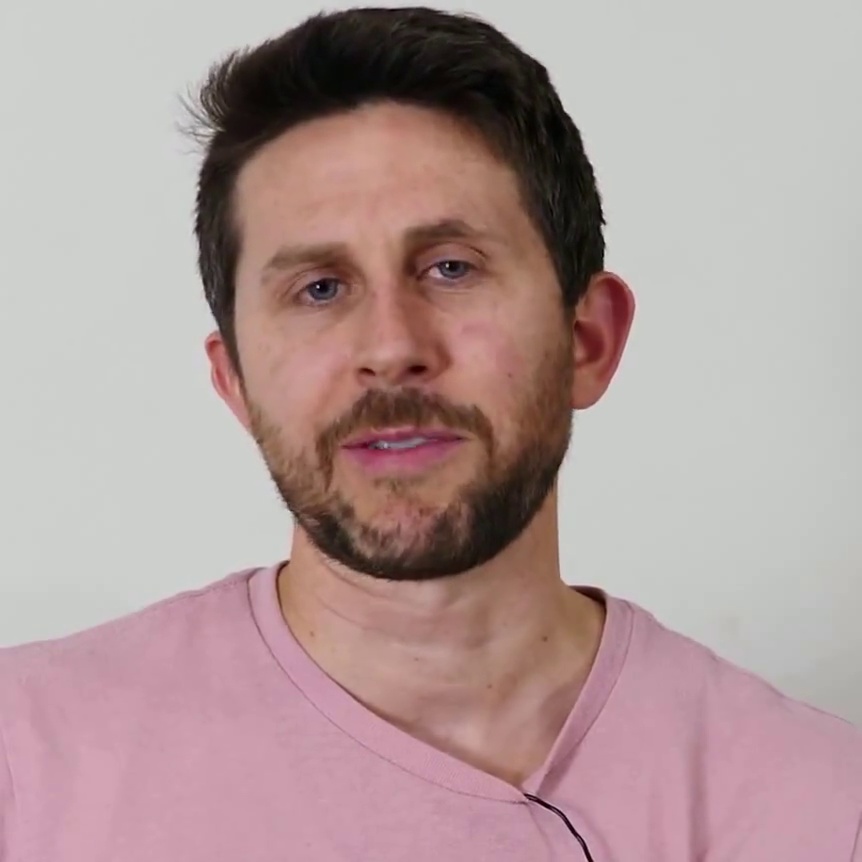} & 
        \includegraphics[width=0.95\linewidth]{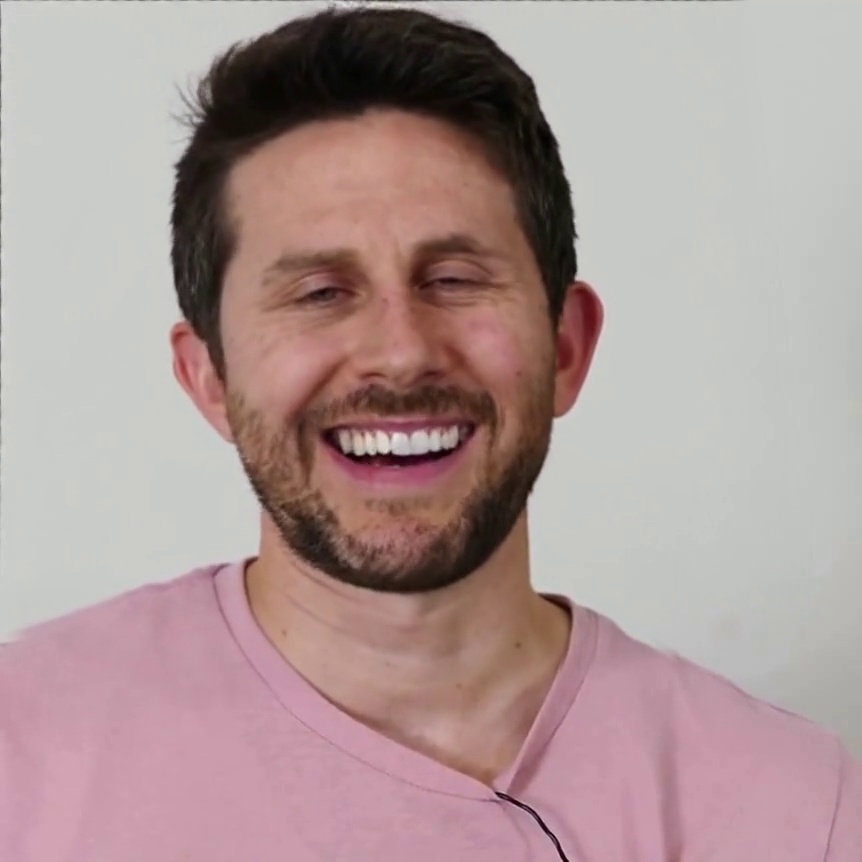} & 
        \includegraphics[width=0.95\linewidth]{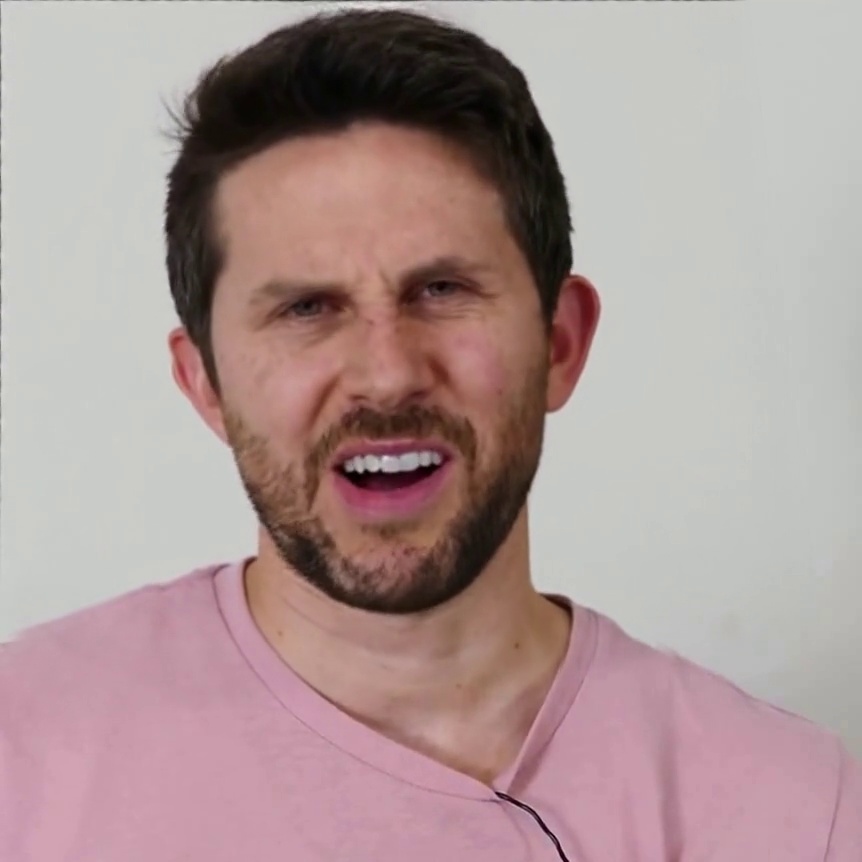} & 
        \includegraphics[width=0.95\linewidth]{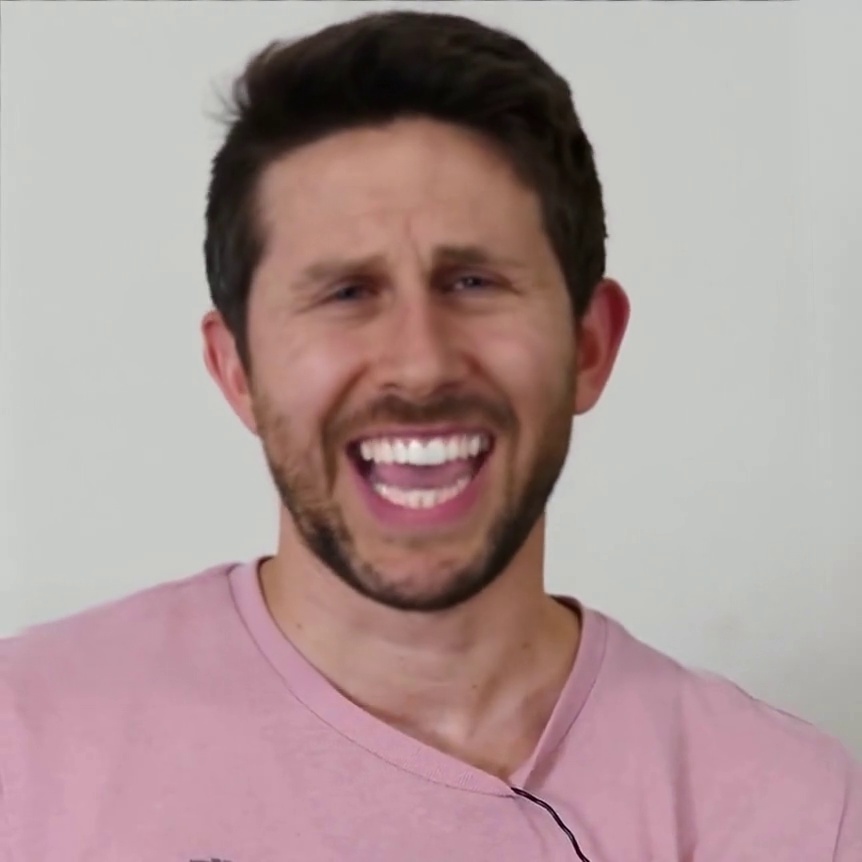} & 
        \textit{\textbf{\footnotesize Smile, Anger, Disgust}}  \includegraphics[width=\linewidth]{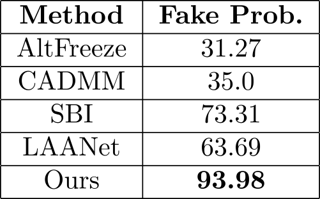} \\       
        \includegraphics[width=0.95\linewidth]{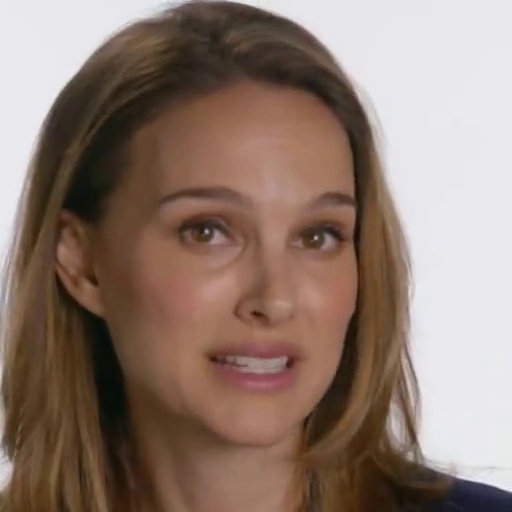} & 
        \includegraphics[width=0.95\linewidth]{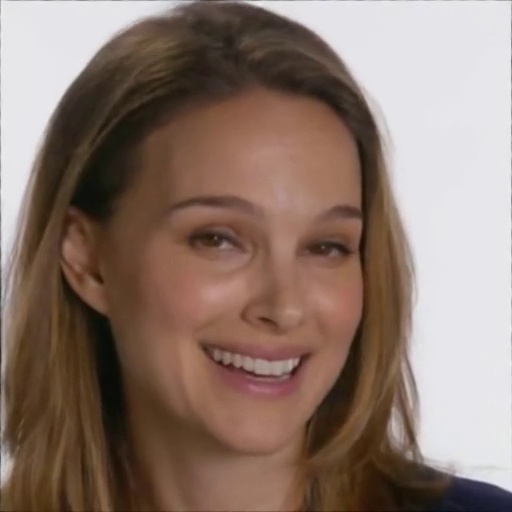} & 
        \includegraphics[width=0.95\linewidth]{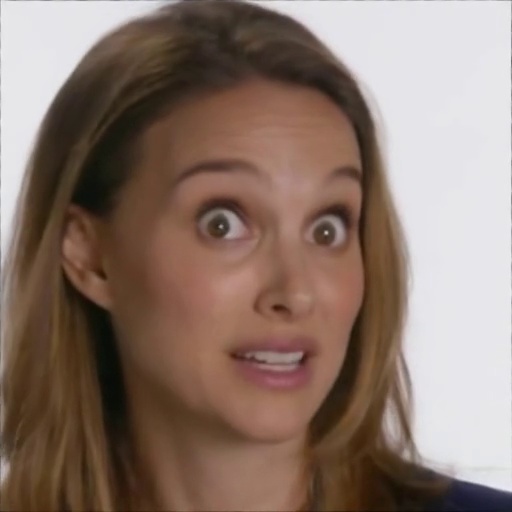} & 
        \includegraphics[width=0.95\linewidth]{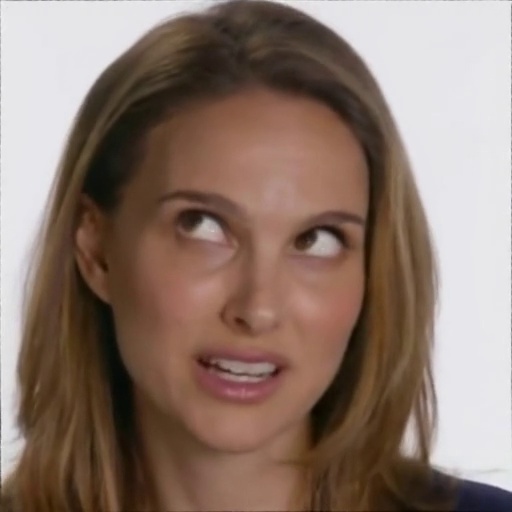} & 
        \textit{\textbf{\footnotesize Smile, EyeRaise, EyeGaze}}  
        \includegraphics[width=\linewidth]{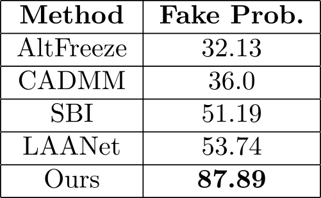} \\               
        \includegraphics[width=0.95\linewidth]{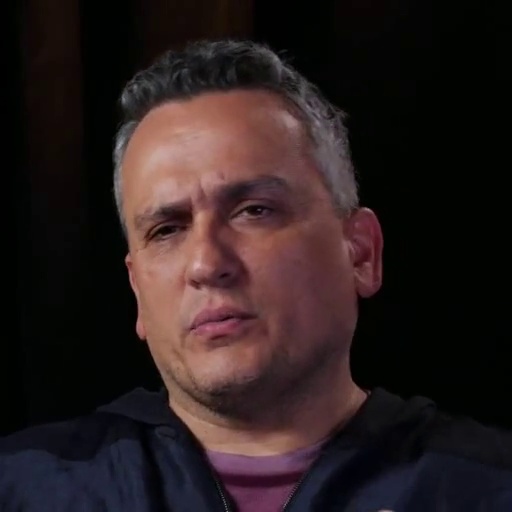} & 
        \includegraphics[width=0.95\linewidth]{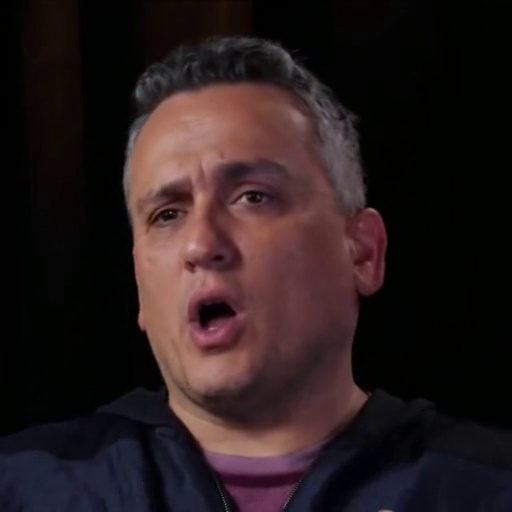} & 
        \includegraphics[width=0.95\linewidth]{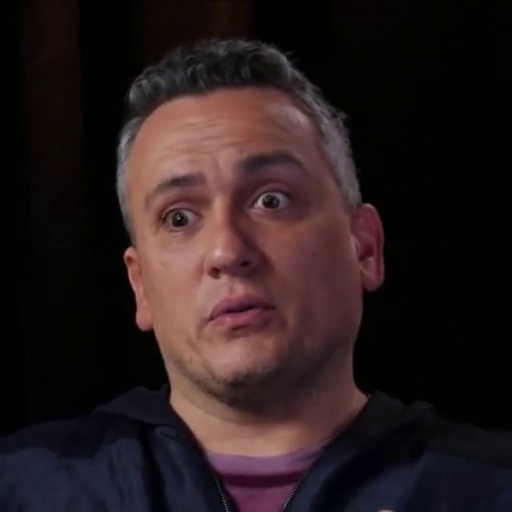} & 
        \includegraphics[width=0.95\linewidth]{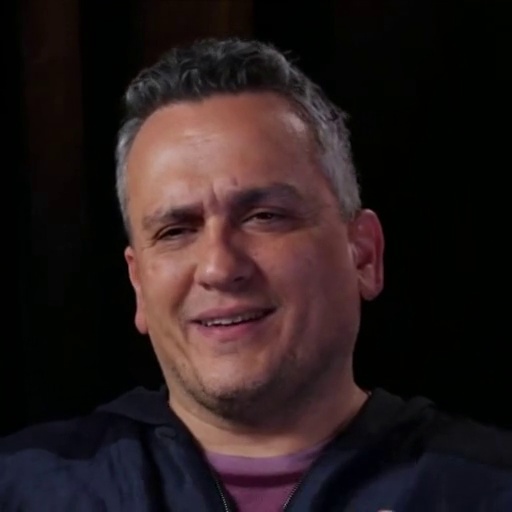} & 
        \textit{\textbf{\footnotesize Shock, EyeRaise, Smile}} 
        \includegraphics[width=\linewidth]{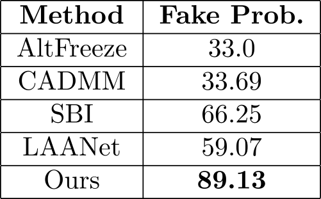} \\        
        \includegraphics[width=0.95\linewidth]{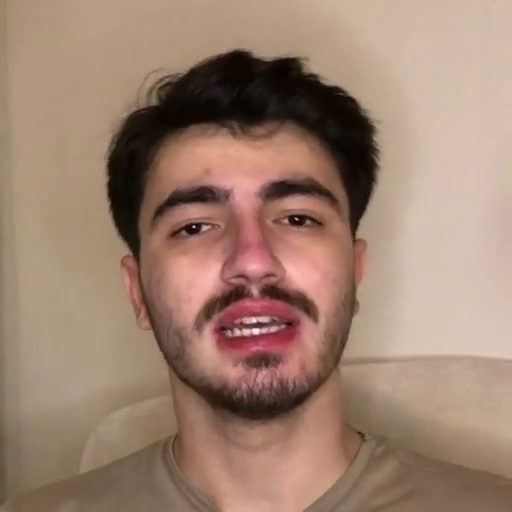} & 
        \includegraphics[width=0.95\linewidth]{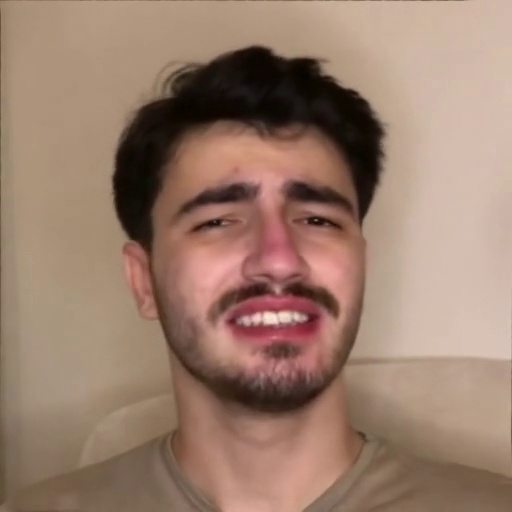} & 
        \includegraphics[width=0.95\linewidth]{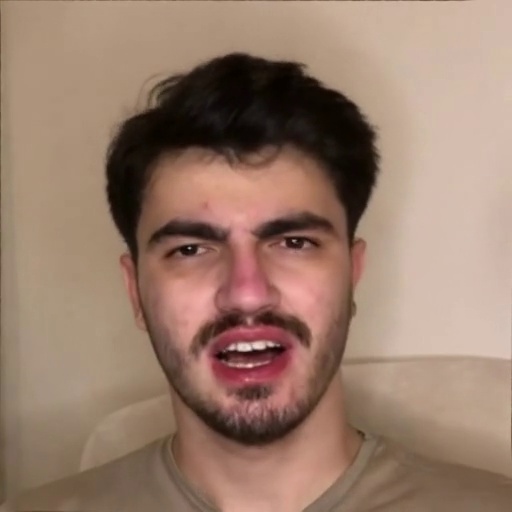} & 
        \includegraphics[width=0.95\linewidth]{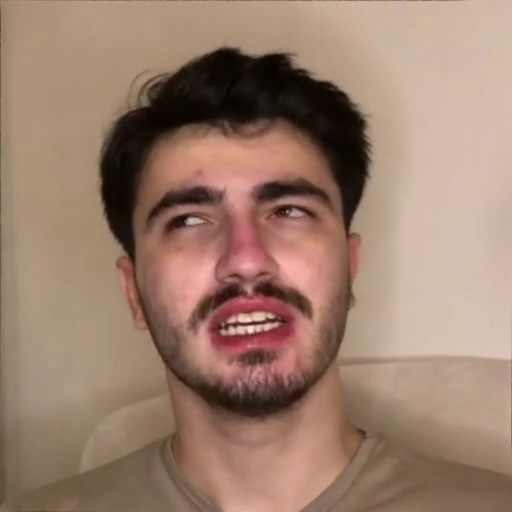} & 
        \textit{\textbf{\footnotesize Disgust, Anger, EyeGaze}} 
        \includegraphics[width=\linewidth]{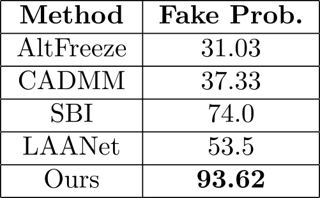} \\ 
        \includegraphics[width=0.95\linewidth]{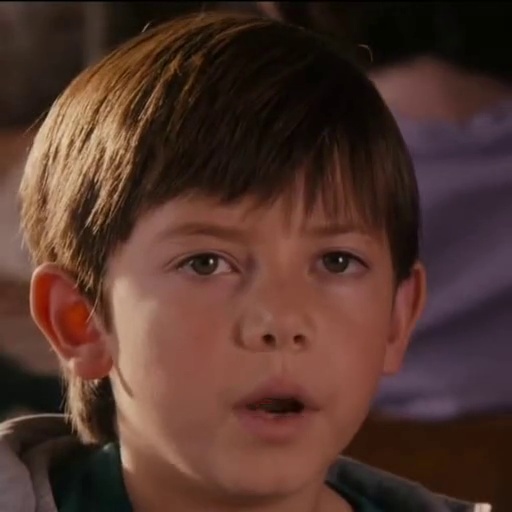} & 
        \includegraphics[width=0.95\linewidth]{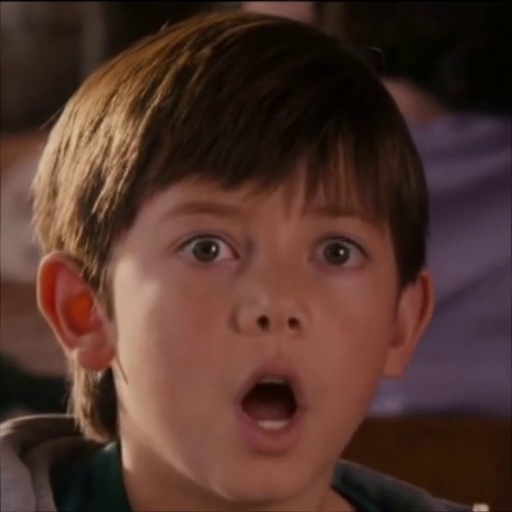} & 
        \includegraphics[width=0.95\linewidth]{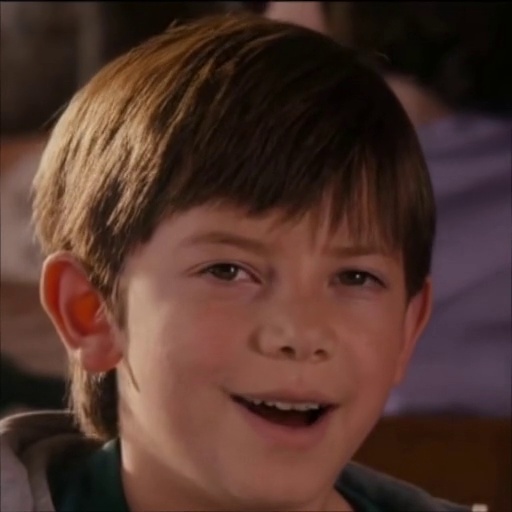} & 
        \includegraphics[width=0.95\linewidth]{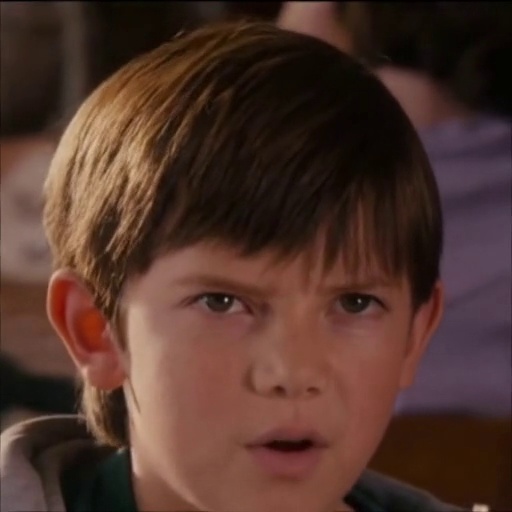} & 
        \textit{\textbf{\footnotesize Shock, Smile, Anger}} 
        \includegraphics[width=\linewidth]{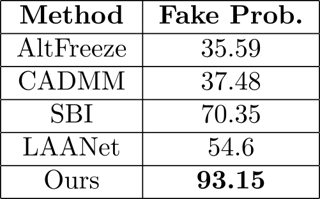} \\
        \textbf{\footnotesize Original} & \multicolumn{3}{c}{\textbf{\footnotesize Fake Manipulations}} & \textbf{\footnotesize \textit{Edit types} \& Probability score}     
    \end{tabular} 
\captionof{figure}{\textbf{Visual Detection Comparison for Locally Manipulated Videos:} A real video (left) undergoes three types of localized manipulations, generating fake videos that are visually indistinguishable from the original. The reported confidence scores, averaged across the three manipulations, highlight our method's superior ability to detect subtle edits compared to the best methods.}
    \label{fig:example}
\end{table*}

\section{Additional Experimental results}
In this section, we evaluate our method's generalization capability in a cross-dataset setting. As shown in Table~\ref{tab:cross_dataset_auc_extended_filled}, our method consistently achieves high performance on standard datasets, exceeding $90\%$ AUC and matching the performance of recent deepfake detection models, LAANet \cite{nguyen2024laa}, SBI \cite{shiohara2023blendface}, AltFreezing \cite{wang2023altfreezing} and CADMM \cite{dong2023implicit}, as shown in Table~\ref{tab:cross_dataset_auc_extended_filled}. In the case of latest locally manipulated video, existing SOTA methods  experience a significant drop in performance. The current SOTA methods exhibit AUCs as low as $30\mbox{-}75\%$, as shown in Table~\ref{tab:cross_dataset_ours}, whereas our method demonstrates robust generalization, achieving an AUC as high as $93\%$. A similar trend is noticeable in the case of all other metrics as well. Notably, our approach exhibits a superior average recall, across all the compared videos, indicating high accuracy in  detecting fake videos (considered as positives), with significantly fewer false negatives and a manageable number of false positives, ensuring efficient and reliable detection even for localized manipulations by recent deepfake methods. 

To visually illustrate our model's superior performance, we display frames of  videos with various localized edits in Fig.~\ref{fig:example}, along with probability scores for detection. All real videos utilized in this experiment are from the publicly available dataset \cite{nagrani2020voxceleb}. Our method consistently achieves confidence scores exceeding $90\%$ in detecting localized edits within fake videos, as compared to the existing state-of-the-art detection methods. This observation holds consistently across a diverse range of localized edits, including expressions such as smiles, shock, disgust, sadness, anger, and modifications like eyebrow raises, eye gaze adjustments, and gender or age transformations. 

Next, in Fig.~\ref{fig:failure_cases}, we present examples where our method exhibits a noticeable drop in confidence scores for detecting fake videos generated through localized edits applied to three real videos in \cite{nagrani2020voxceleb}. Most of these cases occur when the subject is facing sideways or when occlusions hinder the learned representations to accurately capture facial dynamics through action units.

\label{sec:APP_d}
\end{document}